\documentclass[10pt,journal,compsoc]{IEEEtran}
\ifCLASSOPTIONcompsoc
  \usepackage[nocompress]{cite}
\else
  \usepackage{cite}
\fi
\ifCLASSINFOpdf
\else
\fi
\usepackage{epsfig}
\usepackage{graphicx}
\usepackage{amsmath}
\usepackage{amssymb}
\usepackage{subfigure}
\usepackage{bm}
\usepackage[amsmath,thmmarks]{ntheorem}
\usepackage[american]{babel}
\usepackage{microtype}
\usepackage{flushend}

\def\ie{{\emph{i.e.}}}
\def\eg{{\emph{e.g.}}}
\def\etal{{\emph{et al.~}}}

\newtheorem{problem}{Problem}

\def\x{{\mathbf{x}}}
\def\a{{\mathbf{a}}}
\def\e{{\mathbf{e}}}
\def\z{{\mathbf{z}}}
\def\y{{\mathbf{y}}}
\def\s{{\mathbf{s}}}
\def\c{{\mathbf{c}}}
\def\t{{\mathbf{t}}}
\def\K{{\mathbf{K}}}
\def\L{{\mathbf{L}}}
\def\P{{\mathbf{P}}}
\def\I{{\mathbf{I}}}
\def\X{{\mathbf{X}}}
\def\Y{{\mathbf{Y}}}
\def\Q{{\mathbf{Q}}}
\def\A{{\mathbf{A}}}
\def\U{{\mathbf{U}}}
\def\V{{\mathbf{V}}}
\def\S{{\mathbf{S}}}
\def\E{{\mathbf{E}}}

\begin{document}
%
\title{In Defense of Subspace Tracker: Orthogonal Embedding for Visual Tracking}
%
%
%
%

\author{Yao~Sui,~
        Guanghui~Wang,~
        and~Li~Zhang
\IEEEcompsocitemizethanks{\IEEEcompsocthanksitem Y. Sui is with Harvard Medical School, Harvard University, Boston, MA 02115, USA. E-mail: suiyao@gmail.com (Corresponding Author)\protect\\
\IEEEcompsocthanksitem G. Wang is with the Department of Computer Science, Ryerson University, Toronto ON, Canada, M5B 2K3.\protect\\
\IEEEcompsocthanksitem L. Zhang is with the Department of Electronic Engineering, Tsinghua University, Beijing 100084, China.}
\thanks{This work was supported in part by the National Natural Science Foundation of China (NSFC) under Grant 61132007, in part by the Kansas NASA EPSCoR Program under Grant KNEP-PDG-10-2017-KU, in part by the General Research Fund of the University of Kansas under Grant 2228901, and in part by the joint fund of Civil Aviation Research by the National Natural Science Foundation of China (NSFC) and Civil Aviation Administration under Grant U1533132.}
\thanks{Manuscript received May 7, 2018.}}

%
%

\markboth{}%
{Sui \MakeLowercase{\textit{et al.}}: In Defense of Subspace Tracker: Orthogonal Embedding for Visual Tracking}
%



\IEEEtitleabstractindextext{%
\begin{abstract}
The paper focuses on a classical tracking model, subspace learning, grounded on the fact that the targets in successive frames are considered to reside in a low-dimensional subspace or manifold due to the similarity in their appearances. In recent years, a number of subspace trackers have been proposed and obtained impressive results. Inspired by the most recent results that the tracking performance is boosted by the subspace with discrimination capability learned over the recently localized targets and their immediately surrounding background, this work aims at solving such a problem: how to learn a robust low-dimensional subspace to accurately and discriminatively represent these target and background samples. To this end, a discriminative approach, which reliably separates the target from its surrounding background, is injected into the subspace learning by means of joint learning, achieving a dimension-adaptive subspace with superior discrimination capability. The proposed approach is extensively evaluated and compared with the state-of-the-art trackers on four popular tracking benchmarks. The experimental results demonstrate that the proposed tracker performs competitively against its counterparts. In particular, it achieves more than 9\% performance increase compared with the state-of-the-art subspace trackers.
\end{abstract}

}

\maketitle

\IEEEdisplaynontitleabstractindextext

%
\IEEEpeerreviewmaketitle

\IEEEraisesectionheading{\section{Introduction}\label{sec:introduction}}

%
%
%
%
\IEEEPARstart{V}{isual} tracking plays an important role in computer vision with various practical applications like video surveillance, robotics, human-machine interactions, and unmanned control systems. The essential problem to be solved in visual tracking is that, given a motion state of the object target in the initial frame of a video sequence, the visual tracker needs to estimate the subsequent motion states in successive frames. During the past decade, a series of great success in visual tracking has been witnessed \cite{Yilmaz2006,Smeulders2014,sui2016real,bharati2018real,wu2017vision}. There are, however, still many challenges in constructing an effective and robust visual tracker, such as occlusions, illumination changes, background clutters, and non-rigid deformations of the object target.

To deal with these challenges, it is critical for modern tracking algorithms to learn a dynamic target model, also known as the appearance model, to represent the target (and its surrounding background if discriminative information is considered) in successive frames, which can adaptively capture the latest changes of the target in its appearance. One of the most popular methods is the subspace tracking model \cite{Ross2007,Li2007,Kwon2010,Wang2012,Wang2013,Wang2013a,Wang2014,Sui2015tip,Sui2015iccv,Sui2016ijcv,sui2019sparse}. Such popularity is built upon the intuitive and straightforward motivation: the object targets in successive frames are assumed to reside in a low-dimensional subspace and be observed in a high-dimensional space with noise contaminations and corruptions. Note that this assumption is a very weak constraint to visual trackers in a short period of tracking due to the high similarity between the object targets localized in the recent frames. Under this assumption, the object target can be represented accurately and robustly by a low-dimensional subspace, while the distractions are handled by formulating them as an additive error (\ie, noise contaminations and corruptions). In recent literature, the low dimension property has been demonstrated to be effective to challenging situations like pose changes and illumination variations \cite{Hager1996,Kriegmant1996}, and the formulation of additive errors has been proved to be robust against the challenges like occlusions and deformations \cite{Wang2012,Wang2013,Sui2016ijcv}.

The most recent success in subspace tracker \cite{Sui2015iccv} introduced discriminative approach to the subspace tracking model, known as \emph{discriminative subspace learning}. In that study, it is observed that the \emph{recently} obtained targets and their \emph{immediately} surrounding background yield the low-dimensional subspace property. As a result, a locality structure in both temporal and spatial dimensions for the targets and the background, respectively, is exploited by jointly learning a low-dimensional subspace and a linear classifier over those target and background samples, leading to a subspace with greatly improved representation capability. That approach achieves impressive tracking performance, and demonstrates that, by integrating the discriminative cues, subspace learning can be significantly augmented in visual tracking. More importantly, it provides a promising framework to subspace trackers.

Inspired by the previous success, this work aims at improving the performance of subspace trackers with a discriminative framework \cite{Sui2015iccv}. To this end, we address our concerns on discriminative subspace learning for visual tracking as follows.

First of all, how to incorporate discriminative information with subspace learning is critical to the target and background representation. A linear classifier is jointly trained over the subspace \emph{reconstructions} of the target and background samples in \cite{Sui2015iccv}. Note that, instead of subspace \emph{embedding} (representations), the use of reconstructions facilitates the learning method and make it easy to implement, because the discrimination is learned in the same sample space as the low dimension structure; the representation capability, however, may be reduced. On the other hand, from the discriminative perspective, because the reconstructions are unable to reduce any dimension, a large number of training samples are required for the jointly learned linear classifier, known as the \emph{curse of dimensionality} in machine learning \cite{Duda2001}. In visual tracking, however, it is impractical to collect adequate training samples to match the dimension in the original domain, which is usually too high to reach for empirical data. For this reason, a dimension reduction is desired in discriminative subspace learning. In this work, the subspace representations (orthogonal embedding) of the target and background samples with reduced dimensions are leveraged to incorporate with the discriminative cues. The subspace representations are learned over the original target and background samples and their dimension are reduced for better discrimination. A linear classifier is then jointly trained over these dimension-reduced subspace representations, leading to both improved representation and discrimination capabilities.

Another important issue to address in subspace learning is to what degree the dimension should be reduced. It has been shown that higher dimension weakens the generalization capability of the subspace, whereas lower dimension results in less accurate representation. Most of the previous methods employ fixed dimension \cite{Wang2012,Wang2013} or a dimension from a fixed threshold on the differences of the principal components \cite{Duda2001}. However, none of them are able to adopt an appropriate dimension with respect to the different data samples, and in particular for the target and background samples in visual tracking, which often change significantly and drastically in successive frames. In this work, a dimension-adaptive approach to subspace embedding is proposed via a rank minimization method to learn the discriminative subspace. As a result, the dimension is appropriately reduced to ensure good generalization as well as an accurate representation.

In addition, the most obvious weakness of subspace learning is the instability in the presence of outliers, \eg, in the case of occlusions in visual tracking. How to compensate for the weakness of subspace learning is thus crucial to the success of visual tracking. A popular method is to penalize the outliers by additive sparse errors \cite{Wang2012,Zhang2012a,Wang2013,Zhang2012b,Sui2015tip,Sui2016ijcv}, such that the errors are small and sparse for normal samples but large and dense for outliers. Motivated by robust principal component analysis (PCA) \cite{Candes2011}, an additive sparse error term, used to deal with the outliers, is adopted in this work to incorporate with the dimension-adaptive approach in subspace learning, leading to an enhanced robustness of the learned subspace.

Extensive experiments on four standard visual tracking benchmarks are conducted in this work. These experimental results show that the proposed tracker achieves improved competitive performance compared to other state-of-the-art visual trackers. Specifically, it significantly improves the tracking performance by more than 9\% in comparison with the state-of-the-art subspace trackers.

The contributions of this work are addressed as the following three-fold:
\begin{enumerate}
\item we propose a novel subspace learning approach by employing a discriminative embedding for visual tracking, which leads to an effective dimension reduction for both reliable discrimination and accurate representation;
\item we propose a novel formulation to solve the proposed subspace learning by leveraging a Hilbert-Schmidt independence criterion (HSIC), leading to a more robust, dimension-adaptive, and discriminative subspace embedding;
\item the proposed approach improves the tracking performance on four popular benchmarks and yields more than 9\% performance increase over the state-of-the-art subspace trackers.
\end{enumerate}

The remainder of this paper is organized as follows: the related work is reviewed in Section 2; the orthogonal embedding learning is presented in Section 3, which results in a subspace transforming projected coefficients with discrimination capability; the proposed tracking algorithm based on the orthogonal embedding is addressed in Section 4; the experimental results are reported and analyzed in Section 5; Section 6 presents a brief discussion, and the paper is concluded in Section 7.

\section{Related Work}
In this section, we review the work related to ours. We first address the background of the subspace learning in visual tracking, and review state-of-the-art subspace trackers. Besides subspace model, the state-of-the-art tracking frameworks are then presented in brief. The most similar trackers to ours are also  addressed and their main differences are discussed. For comprehensive reviews of tracking algorithms, readers are recommended to refer to \cite{Yilmaz2006,Smeulders2014}. 
\begin{figure*}[t]
  \centering
  \subfigure[reconstruction]{
  \label{fig:illu_dsl}
  \includegraphics[width=0.51\linewidth]{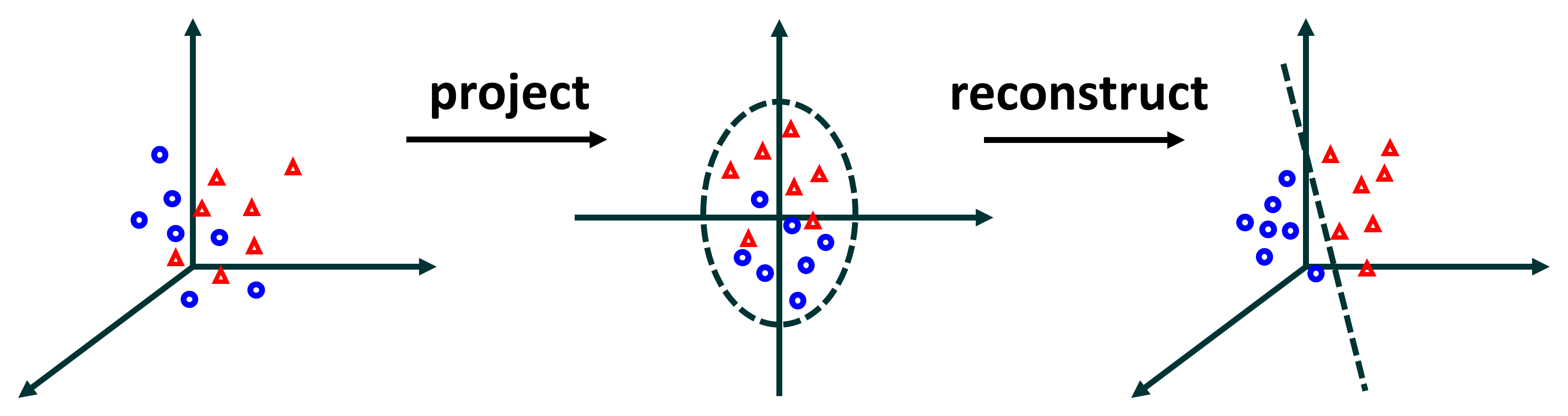}
  }
  \hfil
  \subfigure[embedding]{
  \label{fig:illu_ours}
  \includegraphics[width=0.34\linewidth]{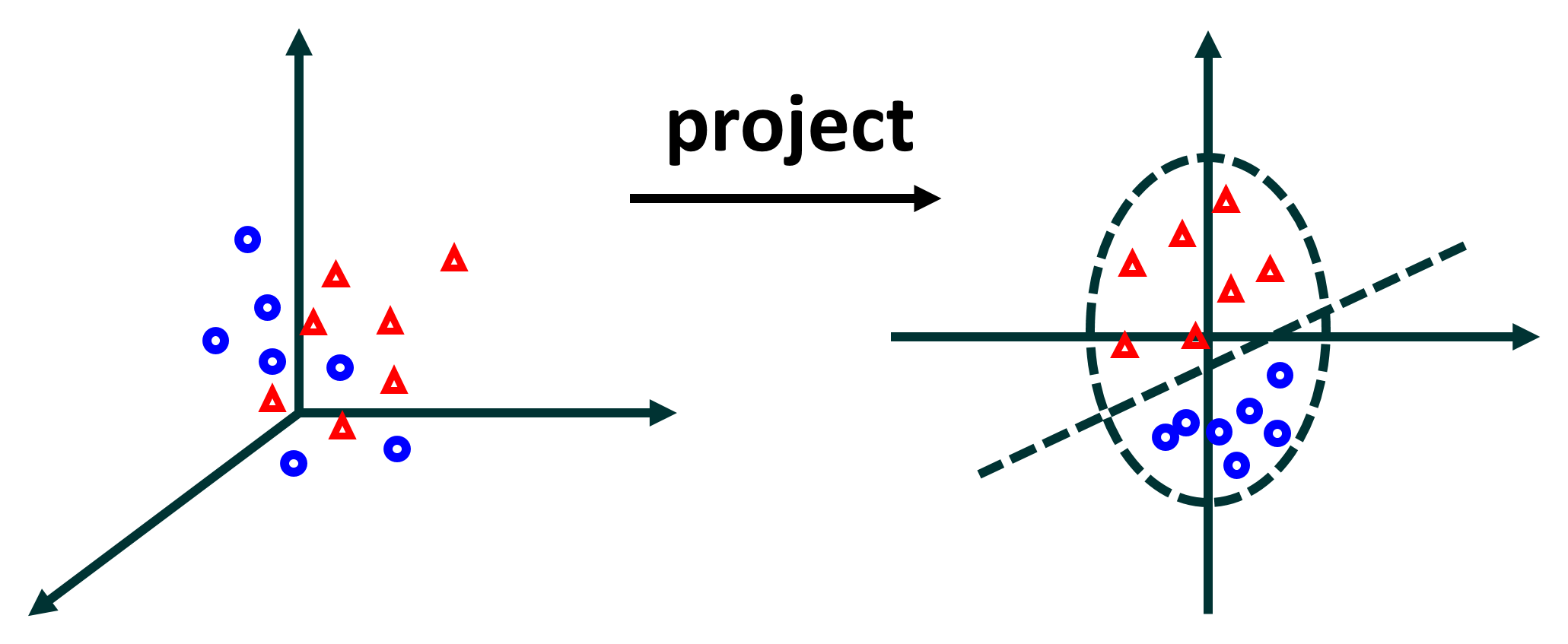}
  }
  \caption{Basic ideas of (a) work \cite{Sui2015iccv} and (b) our approach. The work \cite{Sui2015iccv} leverages subspace \emph{reconstruction} while our approach propagates subspace \emph{embedding} for the discrimination in the joint learning.}
  \label{fig:illu}
\end{figure*}

\subsection{Subspace Trackers}
Subspace learning is a classical method in visual tracking. The basic idea under this paradigm is based on the assumption that the previously obtained targets reside in a low-dimensional subspace in terms of high similarity and strong correlations in their appearances. The underlying assumption behind this paradigm essentially means that the residual errors of the subspace representation yield a Gaussian distribution with a small variance. Such an assumption is effective to the slow and global changes in target appearance, \eg, in the case of illumination variation. However, this assumption also exposes the weakness of subspace learning, \ie, the instability in the presence of outliers, because the residual errors may follow a heavy-tailed distribution with a large variance over the sample sets containing outliers, \eg, in the case of occlusions in visual tracking where target appearances change significantly in successive frames. As a result, most state-of-the-art methods split the residual errors into two parts: a Gaussian (small and dense) error for subspace priors, and a Laplace (large but sparse) error for outliers priors.

Principle component analysis (PCA) method is often used to exploit the subspace priors, which constructs an orthogonal matrix whose columns are considered as the basis vectors of the learned subspace. Eigen-decomposition is a popular algorithm to construct the subspace bases. It leads to dimension reduction when using a few subspace bases corresponding to the first a few largest eigenvalues of sample covariance matrix. A typical tracking model with PCA yields the following steps: 1) compute the basis vectors of the target subspace in the $t$-th frame over $t$ previously obtained target regions; 2) evaluate the representation or reconstruction errors of all candidates in the $\left(t+1\right)$-th frame over the learned target subspace; and 3) determine the candidate with the minimum errors as the $\left(t+1\right)$-th target.

Ross \etal \cite{Ross2007} introduced an incremental algorithm for PCA to visual tracking. Kwon \etal \cite{Kwon2010} employed sparse PCA to decompose visual tracking into different submodules. Wang and Lu \cite{Wang2012} used two-dimensional PCA to construct the subspace in the image domain. Sui \etal \cite{Sui2015tip} proposed a sparsity-induced subspace learning to alleviate the influence of severe occlusions. Wang \etal \cite{Wang2013a} adopted PCA and sparse coding to improve the representation capability of the learned subspace. Zhang \etal \cite{Zhang2012b} proposed a low-rank sparse learning to jointly represent the target and the target candidates. Sui and Zhang \cite{Sui2016ijcv} leveraged a local low-rank and sparse representation to exploit the subspace and neighborhood structures among the local patches within a target region. Wang and Lu \cite{Wang2014} leveraged subspace priors to develop a probability continuous outlier model for visual tracking. Sui \etal \cite{Sui2015iccv} proposed a discriminative subspace learning approach to propagate a linear classifier to a subspace model, obtaining impressive tracking performance. Sui \etal \cite{Sui2018ijcv} imposed a feature evaluation model on the work \cite{Sui2015iccv} to improve tracking accuracy and robustness.

\subsection{Visual Trackers besides Subspace Learning}
Visual tracking algorithms are in general classified into two categories: generative and discriminative. The generative tracking method searches for a target candidate that best matches the current appearance model. Subspace learning is a typical generative tracking approach. Besides that, there are various generative tracking algorithms proposed in recent years, including mean-shift \cite{Comaniciu2003}, sparse learning \cite{Mei2009,Zhang2016}, dictionary learning \cite{Wang2013b}, low-rank learning \cite{Sui2015pr}, and Gaussian process regression \cite{Sui2015spl}. Discriminative tracking method, as known as tracking-by-detection, treats tracking as a binary classification task where the target is separated from its surrounding background. There is extensive literature on discriminative tracking algorithms, such as structural learning \cite{Kalal2012}, multiple instance learning \cite{Babenko2011}, correlation filtering \cite{sui2018joint,Sui2016rcf,Mueller2017,Sui2018tcyb,bharati2016fast}, and deep learning \cite{Huang2015,Qi2016,Nam2016,Choi2017,Han2017,zhang2020real}. Recently, more and more tracking models tend to be both generative and discriminative \cite{Zhong2012,Sui2018tip}, where the tracking model can be generalized over the target regions and distinguish the target from its background reliably.

\subsection{Differences from Similar Trackers}
The work \cite{Sui2015iccv} is the most similar method to our approach. It leverages the subspace \emph{reconstructions} to jointly train a linear classifier, while our approach propagates the subspace \emph{embedding} for the linear classifier with joint learning. The difference of the basic ideas between \cite{Sui2015iccv} and ours is illustrated in Fig. \ref{fig:illu}. It can be seen that \cite{Sui2015iccv} projects the samples onto a low-dimensional subspace and then projects them back to the sample space where the samples are expected to be linearly separable. The dimension of the sample space remains constant for the classifier training, leading to much pressure in the discriminative learning. Different from \cite{Sui2015iccv}, our approach projects the samples onto a low-dimensional subspace and at the same time the projections (\ie, the embedding) are linearly separable. Because the dimension of the embedding space for classifier learning is lower than the training sample space, the discrimination between target and background samples can be learned well even over limited number of samples. In addition to the difference in basic ideas, we create a novel model to formulate the above idea and develop a new approach to solve the proposed problem by introducing the HSIC criterion \cite{Barshan2011}. We demonstrate that using subspace embedding leads to better discrimination capability that is critical to visual tracking.

\section{Orthogonal Embedding}
\subsection{Problem Statement}
Given the recently localized targets and their immediately surrounding background, it is obvious that there are high similarities and strong correlations among these targets and background. It is thus reasonable to assume that these targets and background reside in a low-dimensional subspace and are observed with noise contaminations and corruptions in a high-dimensional space. Therefore, it is desirable to find such a subspace to accurately and robustly represent these targets and background. Meanwhile, the targets are expected to be distinguished from their surrounding background. It indicates that the subspace representations should have discrimination capability over these targets and background. To achieve this goal, we formulate the problem as follows.
\begin{problem}
Construct a dimension-adaptive subspace to represent the recently localized targets and their immediately surrounding background not only accurately and robustly, but also discriminatively.
\end{problem}

Specifically, given $n$ data samples $\left\{\x_i\right\}_{i=1}^{n}$ with $m$ features (\ie, $\x_i\in\mathbb{R}^m$), and the data sample labels $\left\{y_i\right\}_{i=1}^{n}$ that indicate the data samples belong to either \emph{target} or \emph{background}, jointly learn the discriminative mapping function $f\left(\cdot;\bm{\theta}\right)$ defined by the parameters $\bm{\theta}$, and the orthogonal transform function $g\left(\cdot;\bm{\vartheta}\right)$ defined by the parameters $\bm{\vartheta}$,
\begin{equation}
\label{eq:problem}
\begin{split}
&\min_{\a_i,\e_i,\z_i,\bm{\theta},\bm{\vartheta}}
\sum_{i=1}^{n}\ell\left(y_i,f\left(\z_i;\bm{\theta}\right)\right) \\
&s.t.~~
\begin{cases}
\z_i=g\left(\a_i;\bm{\vartheta}\right), \\
\x_i=\a_i+\e_i,~~\left\|\e_i\right\|_0<\varepsilon,
\end{cases}
\end{split}
\end{equation}
with a loss function $\ell\left(\cdot,\cdot\right)$, where $\z_i\in\mathbb{R}^d$ for $d\leq m$ denotes the orthogonal embedding (\ie, subspace representation), and $\x_i$ is decomposed into the noise-free component $\a_i$ and the noise contamination/corruption $\e_i$. The sparsity constraint on $\e_i$ via the $\ell_0$-norm $\left\|\cdot\right\|_0$, which counts the nonzeros of an input vector, against a small number $\varepsilon>0$, treats the distractions in tracking as outliers, like occlusions and local deformations. Note that the key difference between \eqref{eq:problem} and \cite{Sui2015iccv} is that the discrimination is imposed on the subspace representation $\z_i$, rather than on the subspace reconstruction $\a_i$. One advantage of using $\z_i$ is that it allows dimension reduction ($d\leq m$). This advantage can alleviate the pressure of $f\left(\cdot;\bm{\theta}\right)$ from the curse of dimension, leading to significantly improved discrimination capability.

\subsection{Implementation}
The discriminative mapping function $f\left(\z_i;\bm{\theta}\right)$ maps the subspace representation $\z_i$ to the label $y_i$ under the restriction of the loss function $\ell\left(\cdot,\cdot\right)$. Note that $f\left(\z_i;\bm{\theta}\right)$ and $\ell\left(\cdot,\cdot\right)$ essentially establish certain relationship between $\z_i$ and $y_i$ to achieve the discrimination. Various approaches can be used to implement such a relationship, \eg, a least squares classifier interprets a relationship according to mean squared errors, and a mutual correlation criterion describes a relationship in terms of dependency. In this work, we employ a criterion that is very close to the subspace learning problem, the Hilbert-Schmidt independence criterion (HSIC) \cite{Barshan2011}, to implement the relationship between $\z_i$ and $y_i$. The HSIC is in general used to measure the dependence between two random variables. In practice, the empirical HSIC is employed as an efficient estimate. Given the two sets of variables\footnote{Suppose the variables in either set have been centered by their mean.} $\left\{\bm{\alpha}_i\right\}_{i=1}^{n}$ and $\left\{\bm{\beta}_i\right\}_{i=1}^{n}$, the empirical HISC is defined as
\begin{equation}
\label{eq:HSIC}
h=\frac{1}{n-1}tr\left(\K\L\right),
\end{equation}
where $\K$ and $\L$ are kernel matrices for $k_{ij}=\kappa\left(\bm{\alpha}_i,\bm{\alpha}_j\right)$ and $l_{ij}=\tau\left(\bm{\beta}_i,\bm{\beta}_j\right)$ with the kernel functions $\kappa\left(\cdot,\cdot\right)$ and $\tau\left(\cdot,\cdot\right)$, respectively. When \eqref{eq:HSIC} is maximized, the dependence between $\bm{\alpha}_i$ and $\bm{\beta}_i$ is correspondingly maximized. In the problem \eqref{eq:problem}, maximizing the empirical HSIC of the subspace representation $\z_i$ and the corresponding label\footnote{$\y_i$ is reformulated as $\left[1,0\right]^T$ for \emph{target} and $\left[0,1\right]^T$ for \emph{background}.} $\y_i$ can establish a strong dependency between $\z_i$ and $\y_i$, leading to good discrimination capability of the subspace representations $\z_i$.

\subsubsection{Subspace Learning}
The orthogonal transform function $g\left(\cdot,\bm{\vartheta}\right)$ defines the subspace learning in \eqref{eq:problem}. The parameters $\bm{\vartheta}$ are defined by a column-orthogonal matrix $\P\in\mathbb{R}^{m\times d}$ ($d\leq m$), where each column of $\P$ is a basis vector of the learned subspace. As a result, $g\left(\cdot,\bm{\vartheta}\right)$ can be formulated as
\begin{equation}
\label{eq:g}
g\left(\x;\P|\P^T\P=\I\right)=\P^T\x,
\end{equation}
where $\I$ is an identity matrix. It indicates that the orthogonal embedding (subspace representation) can be obtained from $\z_i=\P^T\x_i$ when putting the sparse error $\e_i$ aside. Let $\X\in\mathbb{R}^{m\times n}$ and $\Y\in\mathbb{R}^{2\times n}$ denote the samples $\left\{\x_i\right\}_{i=1}^{n}$ and the labels $\left\{\y_i\right\}_{i=1}^{n}$, respectively. Combining with the discriminative cues established by \eqref{eq:HSIC} in terms of the dependency between $\mathbf{Z}=\P^T\X$ and $\Y$, the empirical HSIC is proportional to
\begin{equation}
\label{eq:tr}
tr\left(\K\L\right)=tr\left(\P^T\X\L\X^T\P\right),
\end{equation}
where a linear kernel function is used for $\K$, \ie, $\K=\X^T\P\P^T\X$. To avoid rank deficiency, a Gaussian kernel is used for $\L$. The basis vectors $\P$ can be found by maximizing the dependence between $\P^T\X$ and $\Y$:
\begin{equation}
\label{eq:P}
\max_\P tr\left(\P^T\X\L\X^T\P\right),
\quad s.t.~\P^T\P=\I.
\end{equation}
Note that the problem in \eqref{eq:P}, also known as the supervised PCA \cite{Barshan2011}, can be solved by applying eigen decomposition on $\X\L\X^T$. The $d$ columns of $\P$ are the $d$ eigenvectors corresponding to the top $d$ eigenvalues of $\X\L\X^T$.

\subsubsection{Dimension Adaptivity}
It is difficult to choose an appropriate value for $d$ when solving \eqref{eq:P}. In this work, we propose an approach, which adaptively assigns a value to $d$ according to the samples $\X$. Because the kernel matrix $\L$ is positive semi-definite, it can be decomposed as $\L=\Q^T\Q$. Then, $\X\L\X^T$ is written as
\begin{equation}
\label{eq:Q}
\X\L\X^T=\X\Q^T\Q\X^T=\A\A^T,
\end{equation}
where $\A=\X\Q^T$. As a result, according to \cite{Candes2011}, \eqref{eq:P} is equivalent to the following rank minimization problem:
\begin{equation}
\label{eq:A}
\min_\A \left\|\A\right\|_*,
\quad s.t.~\left\|\A-\X\Q^T\right\|_F^2\leq\epsilon,
\end{equation}
where $\left\|\cdot\right\|_*$ denotes the nuclear norm that computes the sum of all singular values of an input matrix, $\left\|\cdot\right\|_F$ denotes the \emph{Frobenius}-norm, and $\epsilon>0$ controls the approximation precision.

Minimizing the nuclear norm of $\A$ is equivalent to making the singular values of $\A$ the sparsest. The rank of $\A$ is thus consistent with the number of nonzeros of the singular values. The matrix $\A$ can be decomposed as $\A=\U\S\V^T$ through the singular value decomposition (SVD). Let $r$ denote the rank of $\A$. We have $\A=\U_{m\times r}\S_{r\times r}\V_{n\times r}^T$ and $\U_{m\times r}=\A\V_{n\times r}\S_{r\times r}^{-1}$, where $\S_{r\times r}$ is a diagonal matrix consisting of $r$ non-zero singular values of $\A$. It is easy to show that the solution of the basis vectors $\P$ is the column-orthogonal matrix $\U_{m\times r}$. Correspondingly, the dimension $d$ of the learned subspace is adaptively set to $d=rank\left(\A\right)$.

\subsubsection{Robustness Promotion}
To promote the robustness of the learned subspace against the noise contaminations and corruptions, an additive sparse error term $\E$ is leveraged. The sparsity is used to deal with the distractions in tracking, like occlusions and local deformations. As a result, \eqref{eq:A} is reformulated as
\begin{equation}
\label{eq:E}
\begin{split}
&\min_{\A,\E} \left\|\A\right\|_*+\lambda\left\|\E\right\|_1, \\
&s.t.~\left\|\A-\left(\X-\E\right)\Q^T\right\|_F^2\leq\epsilon,
\end{split}
\end{equation}
where $\lambda>0$ is a weight parameter. The $\ell_1$-norm in \eqref{eq:E} is a convex relaxation of the $\ell_0$-norm, which always results in a sparse solution over empirical data. As discussed above, the basis vectors $\P$ of the learned subspace with an adaptive dimension $d$ is obtained from
\begin{equation}
\label{eq:U}
\P=\A\V_{n\times d}\S_{d\times d}^{-1},
\end{equation}
where $\A=\U\S\V^T$, and $d=rank\left(\A\right)$. As a consequence, the dependency between the subspace representations $\P^T\X$ and the labels $\Y$ is maximized, making the subspace representations $\P^T\X$ discriminative.

\subsection{Optimization}
The problem \eqref{eq:E} is not convex with respect to $\left(\A,\E\right)$. However, it is convex in terms of either $\A$ or $\E$. For this reason, an iterative algorithm can be derived to alternately optimize \eqref{eq:E} over $\A$ and $\E$. Because $\E$ is coupled in the matrix multiplication, a relax variable $\E_2$ is introduced for the convenience of analysis. By renaming $\E$ as $\E_1$, \eqref{eq:E} is reformulated as
\begin{equation}
\label{eq:opt}
\begin{split}
&\min_{\A,\E_1,\E_2} \left\|\A\right\|_*+\lambda\left\|\E_1\right\|_1, \\
&s.t.~\left\|\A-\left(\X-\E_2\right)\Q^T\right\|_F^2\leq\epsilon,~~\E_1=\E_2.
\end{split}
\end{equation}

\noindent\textbf{Compute $\A$}. The problem with respect to $\A$ is defined as
\begin{equation}
\min_{\A} \left\|\A\right\|_*
+\frac{\mu}{2}\left\|\A-\left(\X-\E_2\right)\Q^T\right\|_F^2,
\end{equation}
where $\mu>0$ is a weight parameter instead of $\epsilon$. It can be solved using the singular value shrinkage algorithm \cite{Cai2010}:
\begin{equation}
\A=\U\delta\left(1/\mu,\S\right)\V^T,
\end{equation}
where $\U\S\V^T=\left(\X-\E_2\right)\Q^T$, and $\delta\left(\cdot,\cdot\right)$ is the shrinkage operator defined as
\begin{equation}
\label{eq:shrinkage}
\delta\left(\theta,x\right)=
sign\left(x\right)max\left(0,\left|x\right|-\theta\right),~\theta>0.
\end{equation}

\noindent\textbf{Compute $\E_1$}. The problem with respect to $\E_1$ is defined as
\begin{equation}
\min_{\E_1} \lambda\left\|\E_1\right\|_1
+\frac{\mu}{2}\left\|\E_1-\E_2\right\|_F^2,
\end{equation}
which can be solved using the iterative shrinkage algorithm \cite{Beck2009}:
\begin{equation}
\E_1=\delta\left(\lambda/\mu,\E_2\right).
\end{equation}

\noindent\textbf{Compute $\E_2$}. The problem with respect to $\E_2$ is a least squares problem and has a closed-form solution:
\begin{equation}
\E_2=\left(\A\Q-\X\Q^T\Q+\E_1\right)\left(\Q^T\Q+\I\right)^{-1}.
\end{equation}

The iterative algorithm alternately computes one variable while fixing others until the objective value converges. The convergence is reached when the difference between the objective values at two consecutive iterations is less than a small threshold, \eg, $10^{-8}$ in this work.

\section{Tracking Algorithm}
A region of interest (ROI) in a frame is described by a motion state variable
\begin{equation}
\s=\left\{x,y,\sigma\right\}
\end{equation}
where $\left(x,y\right)$ denotes the 2D position of the ROI and $\sigma$ is the scale coefficient. During tracking, given a motion state $\s$, the corresponding ROI is observed. By cropping out the observed ROI from the frame, and then stacking it into a column vector, a sample is obtained. Note that all the observed ROIs are normalized to the same size, to ensure that all samples have the same length.

In the $j$-th frame, $k$ recently obtained targets are collected, denoted by the matrix $\X^+=\left[\t_{j-k},\t_{j-k+1},\dots,\t_{j-1}\right]$, each column $\t_i$ of which corresponds to the target in the $i$-th frame. The background samples, denoted by the matrix $\X^-$, are collected by shifting the ROIs of $\t_{j-1}$ in the $\left(j-1\right)$-th frame along the horizontal and/or vertical directions by a few pixels. The subspace $\P$ can be learned from \eqref{eq:E} and \eqref{eq:U} over the samples $\X=\left[\X^+,\X^-\right]$.

\begin{figure*}[t]
\begin{center}
  \subfigure[]{
   \label{fig:demo_a}
   \includegraphics[width=0.22\linewidth]{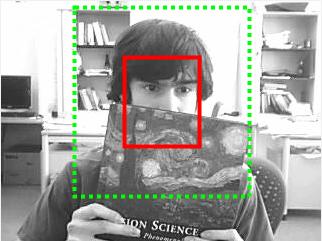}
  }
  \subfigure[]{
   \label{fig:demo_b}
   \includegraphics[width=0.225\linewidth]{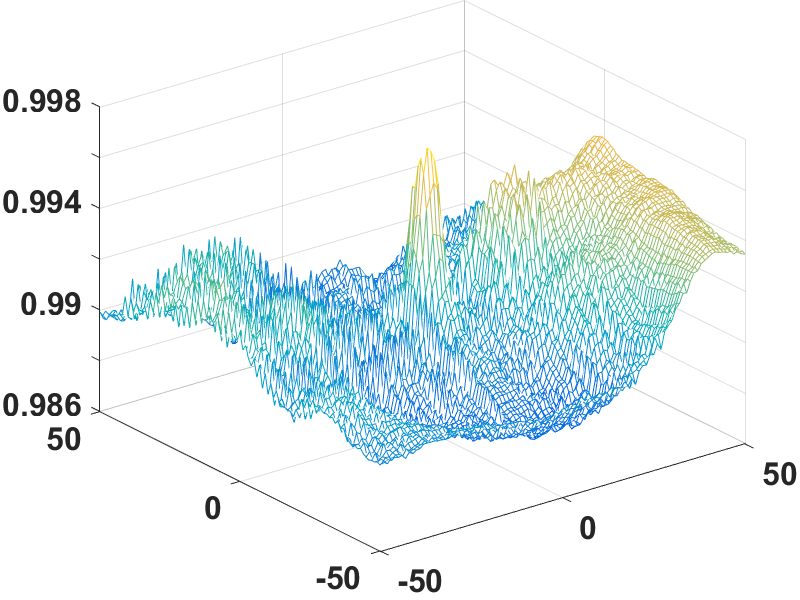}
  }
  \subfigure[]{
   \label{fig:demo_c}
   \includegraphics[width=0.225\linewidth]{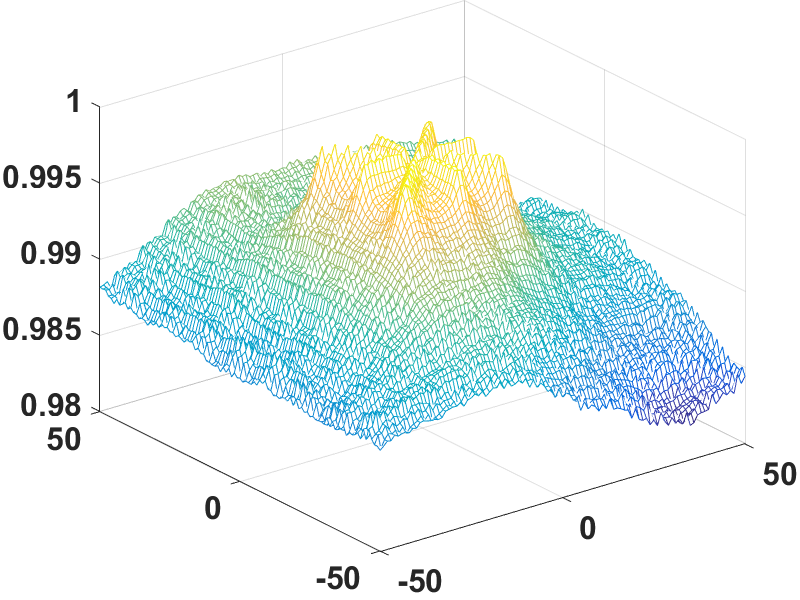}
  }
  \subfigure[]{
   \label{fig:demo_d}
   \includegraphics[width=0.225\linewidth]{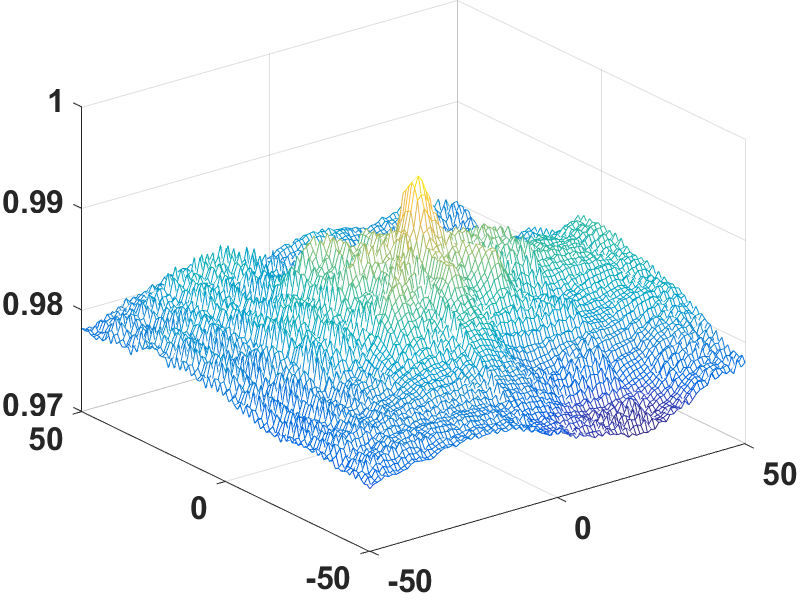}
  }
\end{center}
  \caption{Likelihood to be the target of the candidates within the search area marked by dashed box in a frame shown in (a) in terms of (b) discrimination error $\varphi\left(\cdot\right)$, (c) representation error $\phi\left(\cdot\right)$, and (d) the proposed criterion $\psi\left(\cdot\right)$ defined in \eqref{eq:nt}.}
\label{fig:demo}
\end{figure*}

\subsection{Target Localization}
Suppose a set of target candidates $\mathcal{C}_j$ is available in the $j$-th frame. For a candidate $\c\in\mathcal{C}_j$, its subspace representation $\z$ and residual error $\e$ is obtained from
\begin{equation}
\label{eq:z}
\min_{\z,\e} \left\|\e\right\|_1,
\quad s.t.~~\c=\P\z+\e,
\end{equation}
which can be solved by an iterative algorithm similar to that for \eqref{eq:E}. Note that the error $\e$ indicates the representation accuracy of the learned subspace $\P$ for the candidate $\c$. A criterion, called the representation error in this work, is thus defined as
\begin{equation}
\label{eq:rs}
\phi\left(\c\right)=\left\|\e\right\|_1,
\end{equation}
for a candidate $\c$. The candidate with higher accuracy (\ie, smaller $\phi\left(\c\right)$) is more likely to be the $j$-th target.

Furthermore, the subspace representation $\z$ reflects that the candidate $\c$ is either a target or a background. The discriminative information contained in $\z$ can be measured by the empirical HSIC from \eqref{eq:HSIC} and the target/background labels. The candidate $\c$ is considered as either a target or a background according to which label leads to a larger empirical HSIC with $\z$. Because the labels used in the empirical HSIC are vectors, to efficiently obtain the classification of a candidate, a linear classifier using scalar labels, such as least squares classifier and SVM (our choice), is trained over $\z$ in the target localization. It will be shown in the following that the linear classifiers tend to output consistent results with the empirical HSIC in the target localization. As a result, another criterion, called the discrimination error in this work, is defined as
\begin{equation}
\varphi\left(\c\right)=\left|\pi\left(\z\right)-1\right|,
\end{equation}
for a candidate $\c$. The candidate whose classification response $\pi\left(\z\right)$ is closer to the target label is more likely to be the $j$-th target.

From the above discussion, the target in the $j$-th frame can be localized from a set of candidates $\mathcal{C}_j$ by minimizing the following criterion,
\begin{equation}
\label{eq:t}
\psi\left(\c\right)=\phi\left(\c\right)+\lambda\varphi\left(\c\right).
\end{equation}
Note that, in \eqref{eq:t}, the weight $\lambda>0$ is difficult to tune. Referring to \cite{Sui2015iccv}, the normalized version of above criterion is adopted to avoid tuning $\lambda$:
\begin{equation}
\label{eq:nt}
\psi\left(\c\right)=\frac{\phi\left(\c\right)}{\left\|\left[\phi\left(\c_i\right)\right]_{\c_i\in\mathcal{C}_j}\right\|_2}
+\frac{\varphi\left(\c\right)}{\left\|\left[\varphi\left(\c_i\right)\right]_{\c_i\in\mathcal{C}_j}\right\|_2}.
\end{equation}

\subsection{Tracking Framework}
In the $j$-th frame, given all previously obtained target samples $\t_1,\t_2,\dots,\t_{j-1}$, the motion state $\s_j$ of the $j$-th target can be predicted by
\begin{equation}
\label{eq:s}
p\left(\s_j|\t_{1:j-1}\right)=\int p\left(\s_j|\s_{j-1}\right)p\left(\s_{j-1}|\t_{1:j-1}\right)d\s_{j-1},
\end{equation}
where $p\left(\s_j|\s_{j-1}\right)$ denotes the motion model, which is related to the candidate generation. Then, the corresponding candidate $\c$ is observed and \eqref{eq:s} is updated by
\begin{equation}
\label{eq:c}
p\left(\s_j|\c,\t_{1:j-1}\right)=\frac{p\left(\c|\s_j\right)p\left(\s_j|\t_{1:j-1}\right)}{p\left(\c|\t_{1:j-1}\right)},
\end{equation}
where $p\left(\c|\s_j\right)$ denotes the observation model. The $j$-th target is then found by
\begin{equation}
\label{eq:target_location}
\t_j=\arg\max_{\c\in\mathcal{C}_j}p\left(\s_j|\c,\t_{1:j-1}\right).
\end{equation}

In this work, the motion model is set by a Gaussian distribution $p\left(\s_j|\s_{j-1}\right)\sim\mathcal{N}\left(\mathbf{0},\bm{\Sigma}\right)$ where the diagonal matrix $\bm{\Sigma}$ denotes the covariance of the motion states. The observation model $p\left(\c|\s_j\right)$ is set to be inversely proportional to \eqref{eq:nt}.

\subsection{Update Scheme}
As the target appearance varies over frames, the subspace needs to be updated dynamically over the temporally localized target and the background samples to capture their latest changes. In this work, a buffer is used to maintain the recently localized targets. In each frame, the target stored in the buffer for the longest time is popped out and the latest target is appended to the buffer. Because the target in the initial frame is known as ground truth, it is always kept in the buffer. The background samples are updated in each frame by the immediately surrounding samples of the latest target. Considering the computational efficiency of the proposed tracker, the subspace is re-learned every ten frames over the dynamically maintained target and the background samples.

\subsection{Discussions}
\begin{figure}[t]
\begin{center}
  \subfigure[]{
   \label{fig:hsic_a}
   \includegraphics[width=0.45\linewidth]{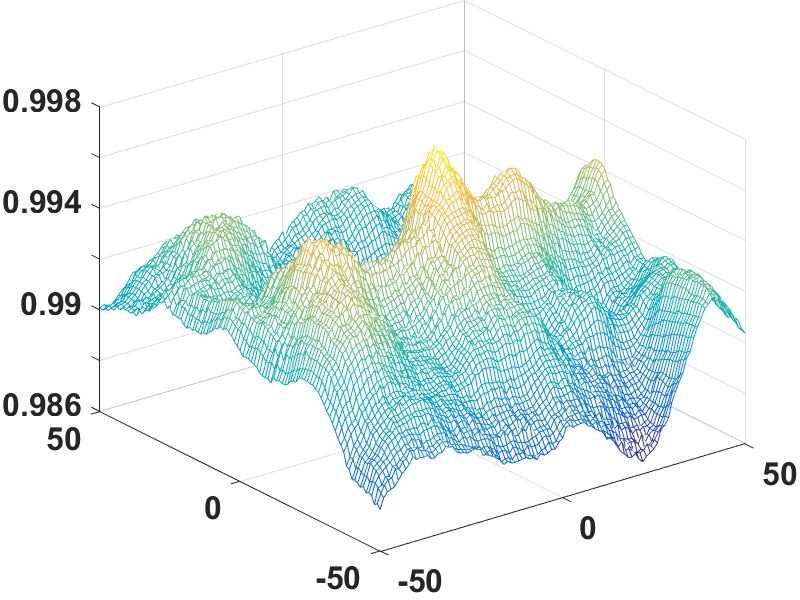}
  }
  \subfigure[]{
   \label{fig:hsic_b}
   \includegraphics[width=0.45\linewidth]{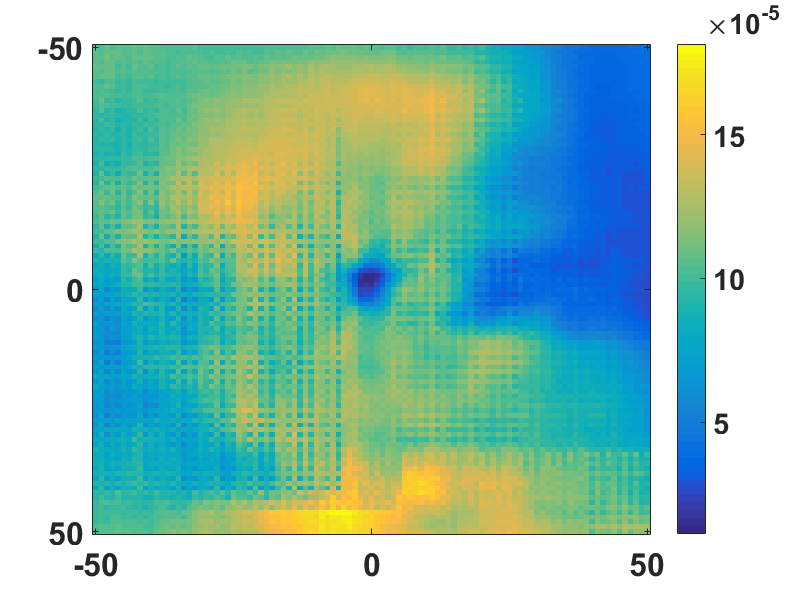}
  }
\end{center}
  \caption{(a) Likelihood of the candidates in Fig. \ref{fig:demo_a} in terms of the empirical HSIC. (b) The difference between the likelihood obtained from the empirical HSIC and the linear classifier.}
\label{fig:hsic}
\end{figure}
\begin{figure*}[t]
\begin{center}
\subfigure[]{
   \label{fig:se_case_a}
  \includegraphics[width=0.225\linewidth]{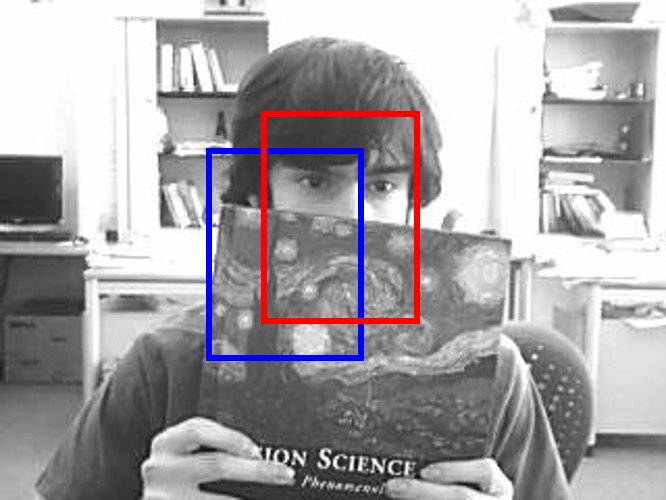}
  }
  \hfil
  \subfigure[]{
   \label{fig:se_case_b}
  \includegraphics[width=0.225\linewidth]{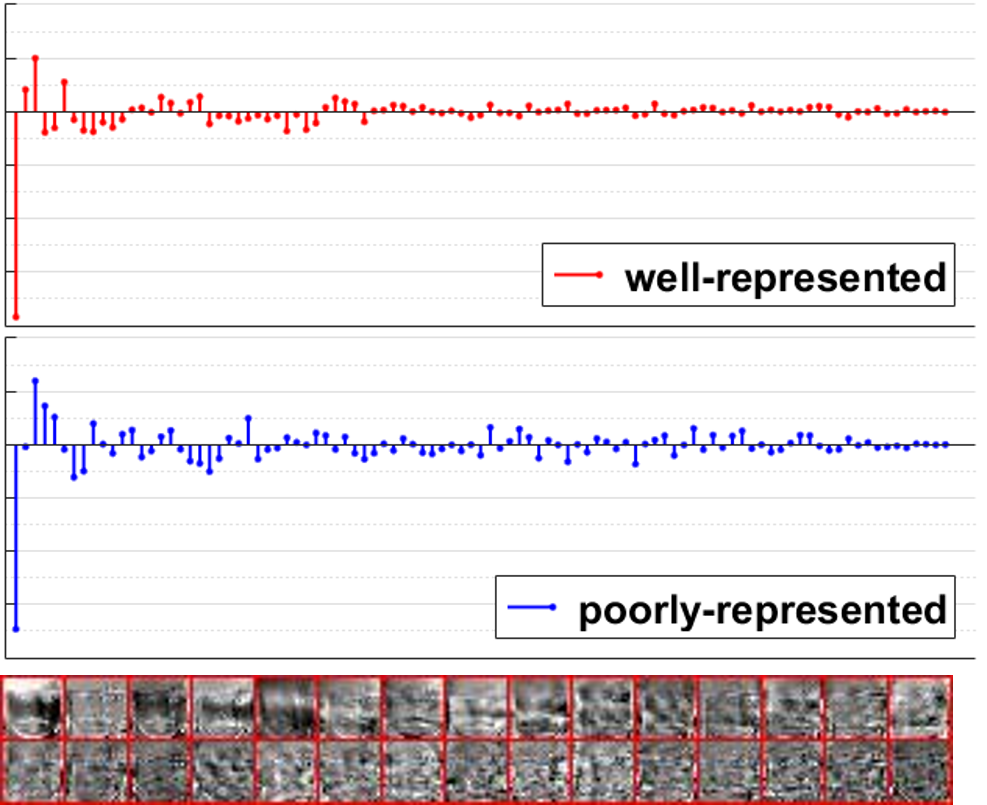}
  }
  \hfil
  \subfigure[]{
   \label{fig:se_case_c}
  \includegraphics[width=0.225\linewidth]{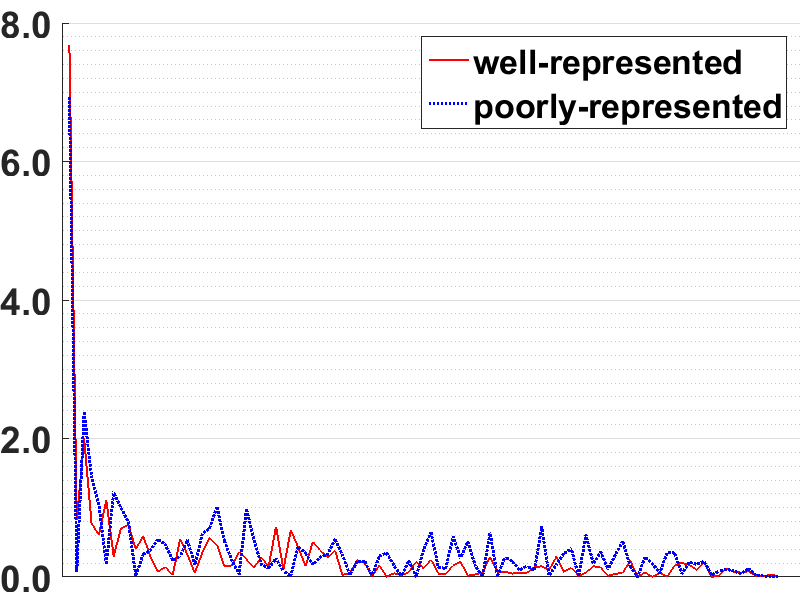}
  }
  \hfil
  \subfigure[]{
   \label{fig:se_case_d}
  \hfil
  \includegraphics[width=0.225\linewidth]{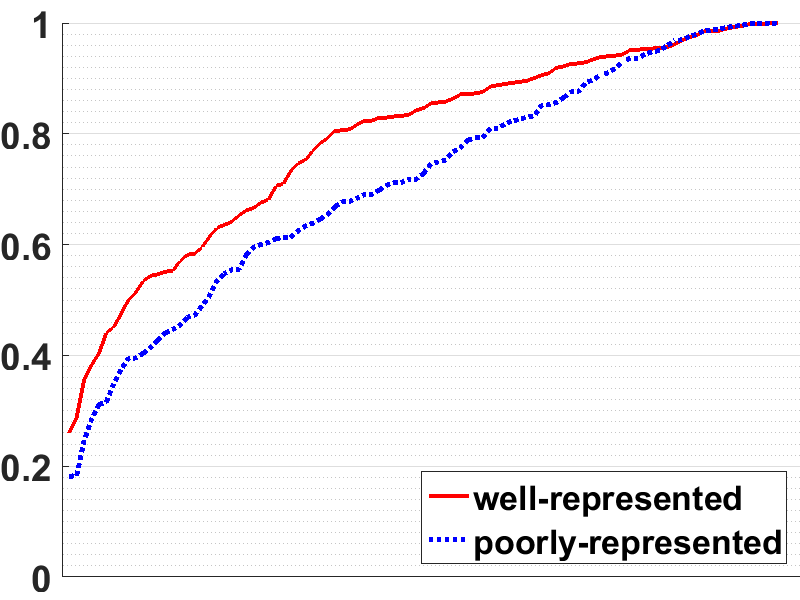}
  }
\end{center}
  \caption{Subspace embedding for a well- and a poorly-represented candidate in a representative frame. (a) A frame where two candidates are highlighted (red for the well- and blue for the poorly-represented); (b) embedding of the two candidates in each dimension over the subspace bases shown in the bottom; (c) magnitudes of the embedding for the two candidates; and (d) cumulative summation curves of the embedding magnitudes for the two candidates.}
  \label{fig:se_case}
\end{figure*}
To demonstrate the effectiveness of the proposed target localization criterion, an example in the case of occlusions is shown as follows. Fig. \ref{fig:demo_a} shows a frame where the target is marked in red box, and a search area centering at the target location is marked by the dashed box. The likelihood to be the target of the candidates centering at each pixel within the search area is investigated in terms of different criteria. The maximum likelihood is expected to appear at the center of the likelihood map. As shown in Fig. \ref{fig:demo_b}, because the background samples are collected from the regions close to the target location, the discrimination error criterion may be unstable in the regions far away from the target location due to the lack of training samples, but it is effective enough to separate the target from its nearby background. On the other hand, as shown in Fig. \ref{fig:demo_c}, the representation error criterion is unable to distinguish the target from its nearby background, but it is effective to recognize the spatially distant background, because the learned subspace represents both the target and the nearby background more accurately than the background located farther away. As a result, the proposed criterion defined in \eqref{eq:nt} enhances the discrimination in the nearby regions and punishes the background at distance, as shown in Fig. \ref{fig:demo_d}.

Moreover, the likelihood in terms of the empirical HSIC is also shown in Fig. \ref{fig:hsic_a}. Although the empirical HSIC does not lead to the exactly same likelihood as the discrimination error in this example, the two criteria tend to produce consistent results. The difference, measured by squared errors, between the likelihood obtained from the two criteria is subsequently investigated, as shown in Fig. \ref{fig:hsic_b}. It is evident that the differences are very small, and the minimum, maximum, and mean differences are $1.1\times10^{-5}$, $1.8\times10^{-4}$, and $9.7\times10^{-5}$, respectively. It indicates that the linear classifier tends to yield consistent results with the empirical HSIC. However, the linear classifier is more efficient to obtain the classification results.

We show another example to discuss the effectiveness of the proposed subspace embedding. We take a representative frame and investigate two candidates, as shown in Fig. \ref{fig:se_case_a}, where a well- and a poorly-represented candidate (\ie, two candidates with respectively more and less likely to be the target) are highlighted in the red and blue boxes, respectively. Fig. \ref{fig:se_case_b} shows the proposed embedding (\ie, projection coefficients in the learned subspace) of the two candidates, and the subspace bases in the bottom. In Figs. \ref{fig:se_case_c} and \ref{fig:se_case_d}, the magnitudes of the embedding and their cumulative summation curves for the two candidates are plotted respectively. It can be seen that the well-represented candidate has large embedding values in the first several dimensions while very small values in the remaining dimensions. It indicates that the well-represented candidate can be described well by only a few bases corresponding to the first a few largest eigenvalues. The poorly-represented candidate, in contrast, has large embedding values in almost all the dimensions. It means that much more subspace bases are required to make a good representation for this poorly-represented candidate. From this example, we can see that the proposed embedding leads to effective dimension reduction, and performs discriminatively to distinguish the target from its background.

\begin{figure*}[t]
\begin{center}
   \includegraphics[width=0.45\linewidth]{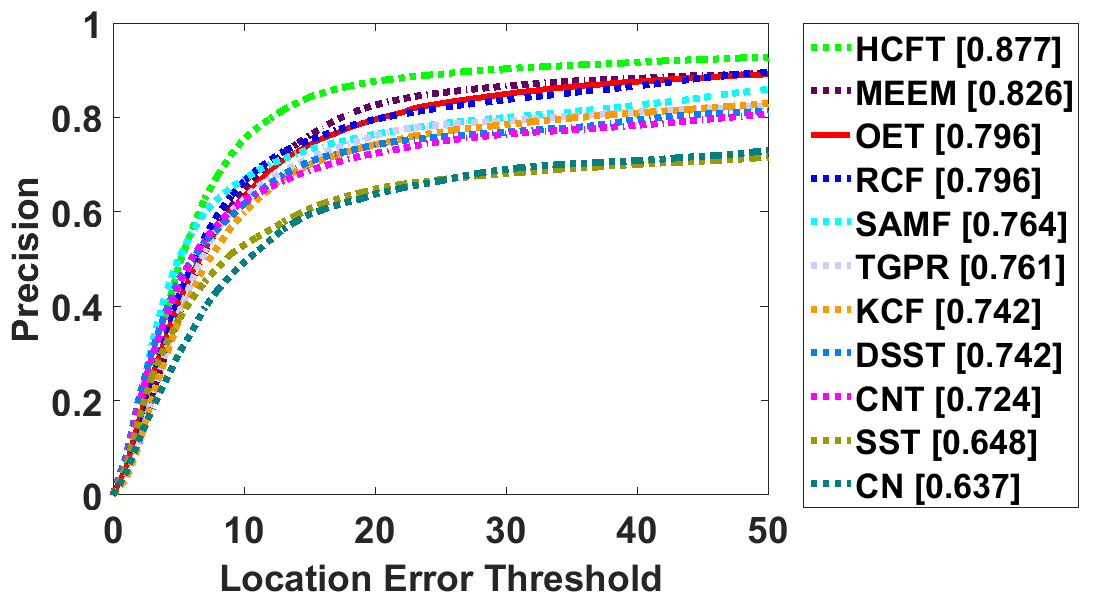}\hfil
   \includegraphics[width=0.45\linewidth]{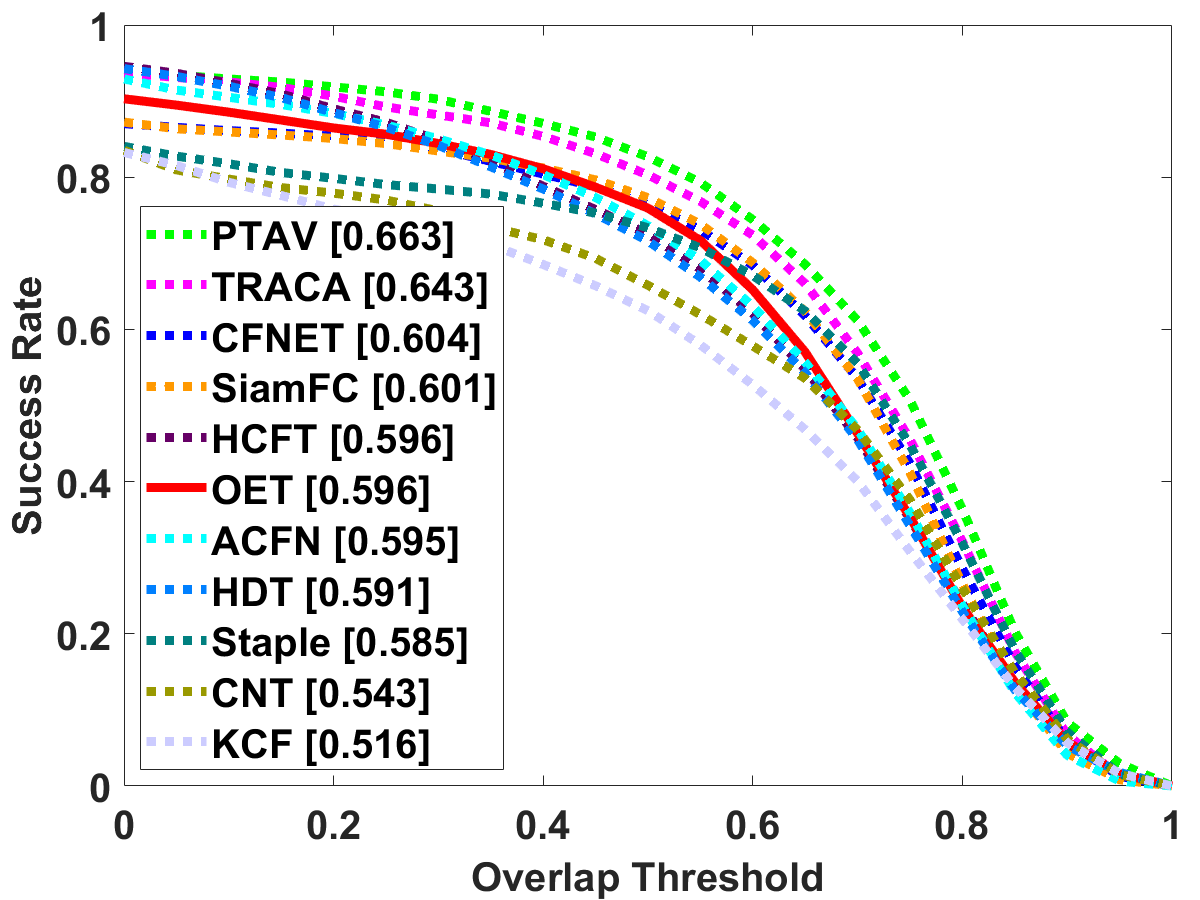}
\end{center}
  \caption{Tracking performance of the proposed tracker compared with 10 state-of-the-art trackers on the OTB-50 benchmark. In the legend of the left figure, the precision at threshold 20 is reported, and in the legend of the right figure, the area under curve (AUC) of the success rate is reported.}
\label{fig:result}
\end{figure*}
\begin{figure*}[t]
\begin{center}
   \includegraphics[width=0.45\linewidth]{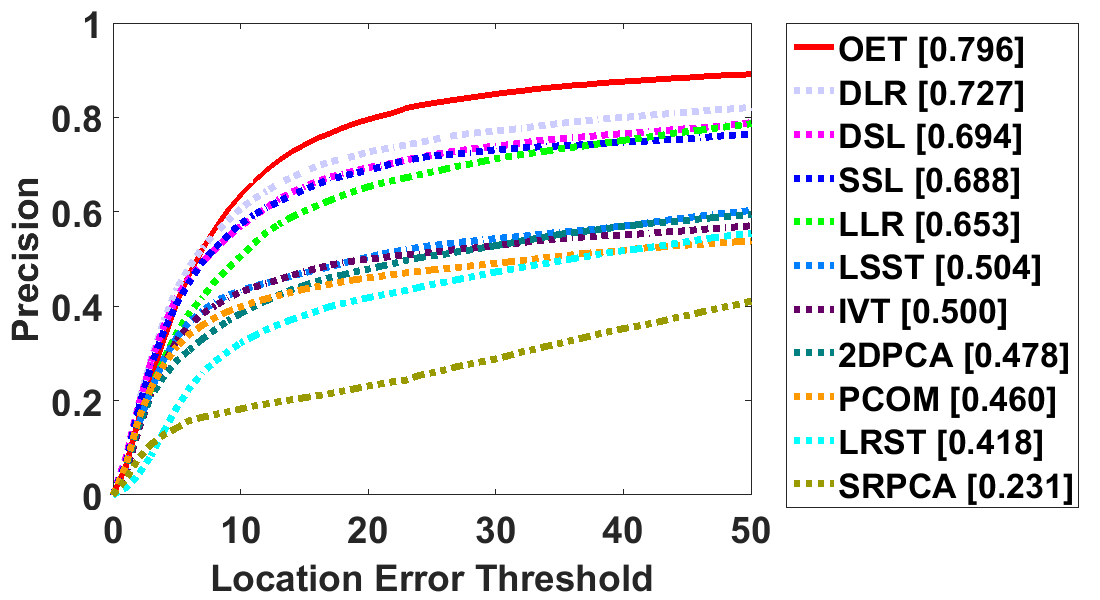}\hfil
   \includegraphics[width=0.45\linewidth]{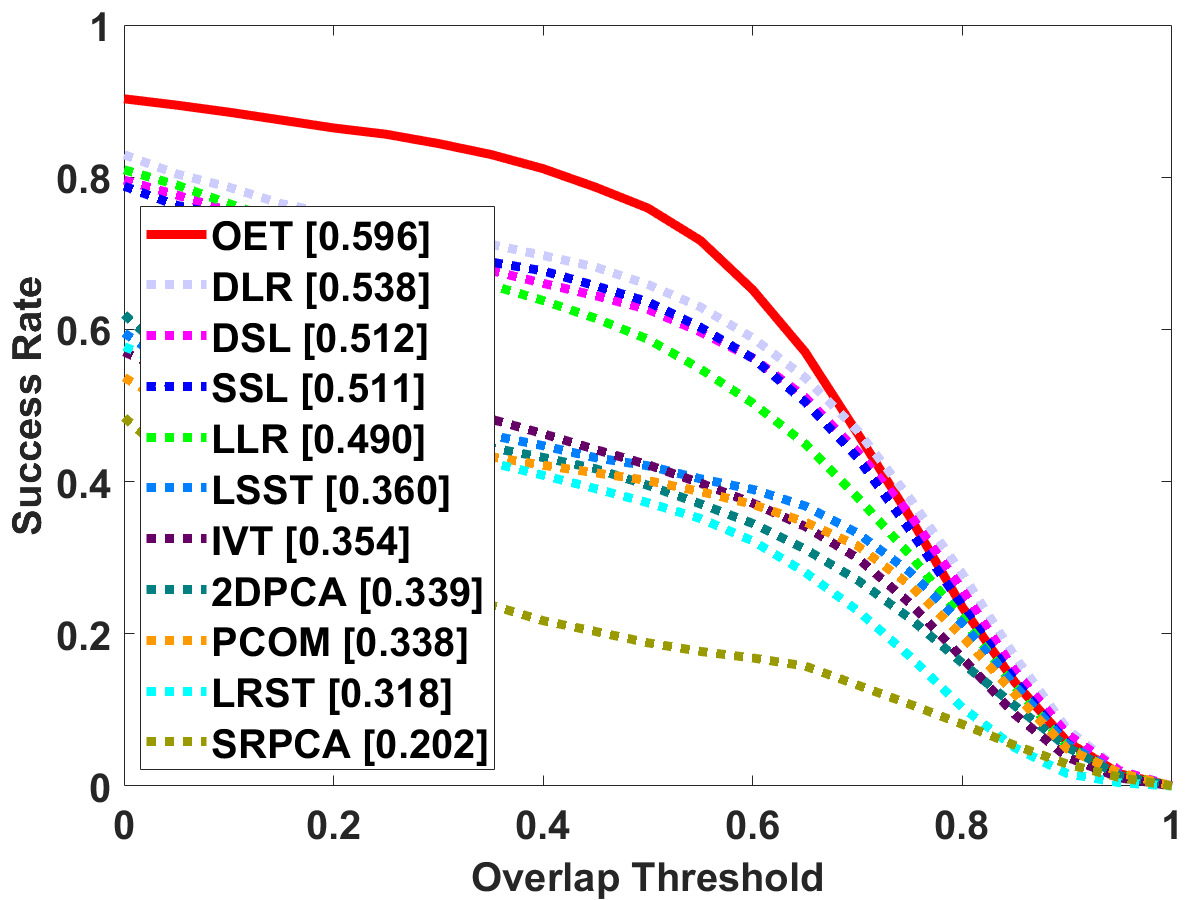}
\end{center}
  \caption{Tracking performance of the proposed tracker compared with 10 latest subspace trackers on the OTB-50 benchmark. In the legend of the left figure, the precision at threshold 20 is reported, and in the legend of the right figure, the area under curve (AUC) of the success rate is reported.}
\label{fig:result_s}
\end{figure*}
\begin{figure*}[t]
\begin{center}
   \includegraphics[width=0.45\linewidth]{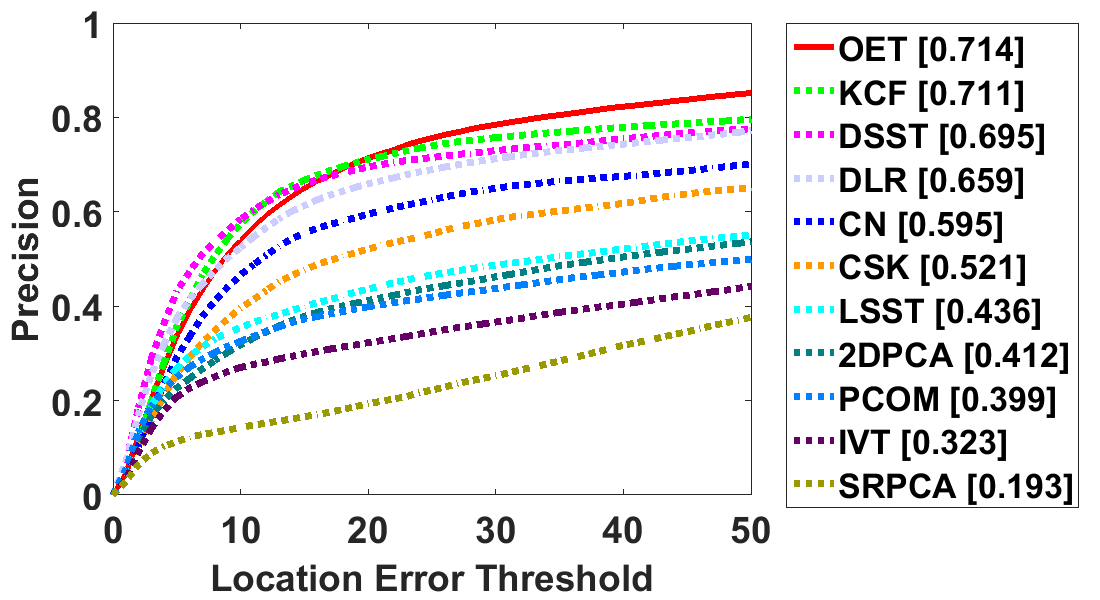}\hfil
   \includegraphics[width=0.45\linewidth]{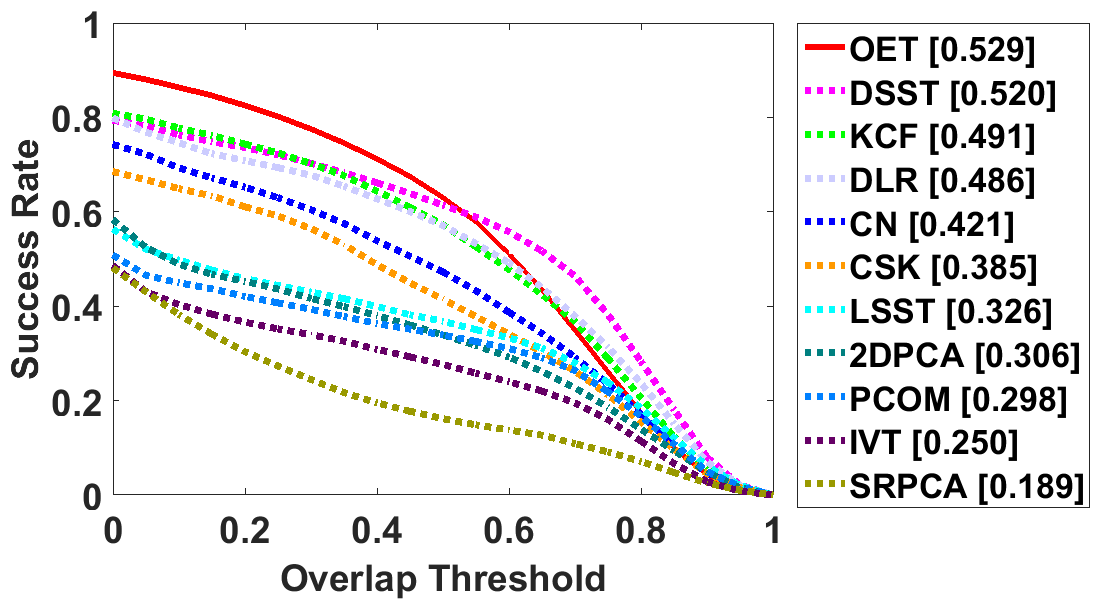}
\end{center}
  \caption{Tracking performance of the proposed tracker compared with other 10 state-of-the-art trackers on the OTB-100 benchmark. In the legend of the left figure, the precision at threshold 20 is reported, and in the legend of the right figure, the area under curve (AUC) of the success rate is reported.}
\label{fig:result_otb100}
\end{figure*}
\begin{figure*}[t]
\begin{center}
  \subfigure[Occlusions]{
   \includegraphics[width=0.225\linewidth]{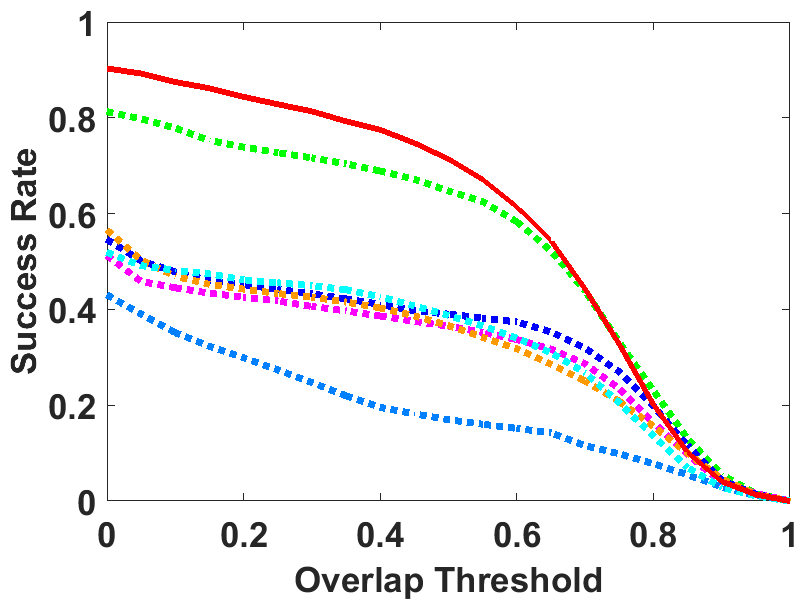}
  }\hfil
  \subfigure[Illumination variations]{
   \includegraphics[width=0.225\linewidth]{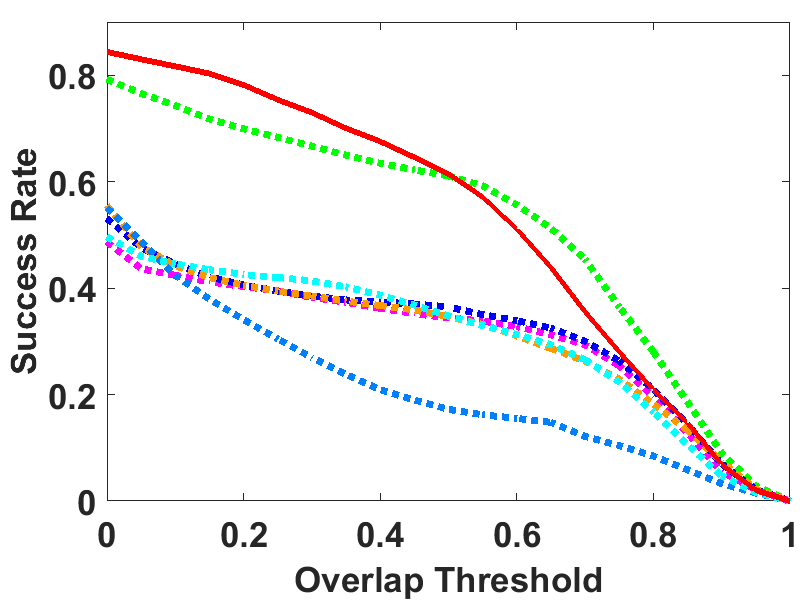}
  }\hfil
  \subfigure[Background clutters]{
   \includegraphics[width=0.225\linewidth]{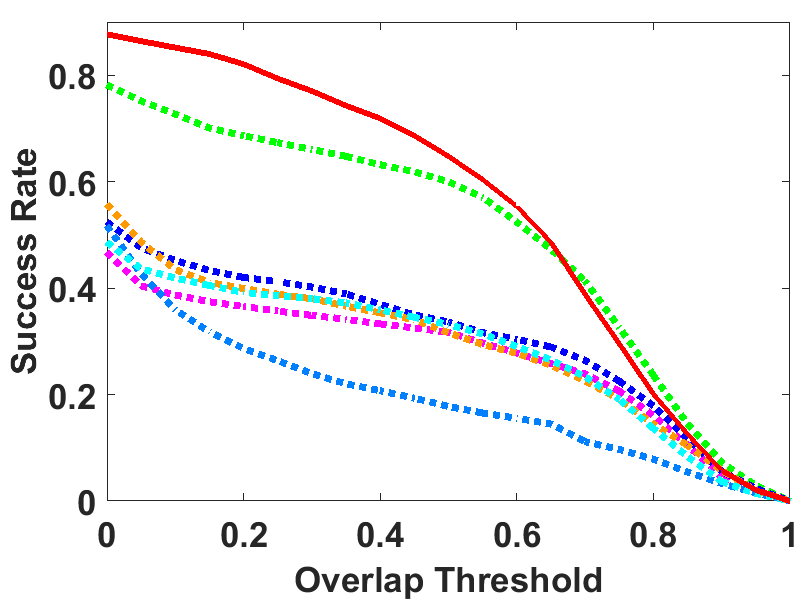}
  }\hfil
  \subfigure[Non-rigid deformations]{
   \includegraphics[width=0.225\linewidth]{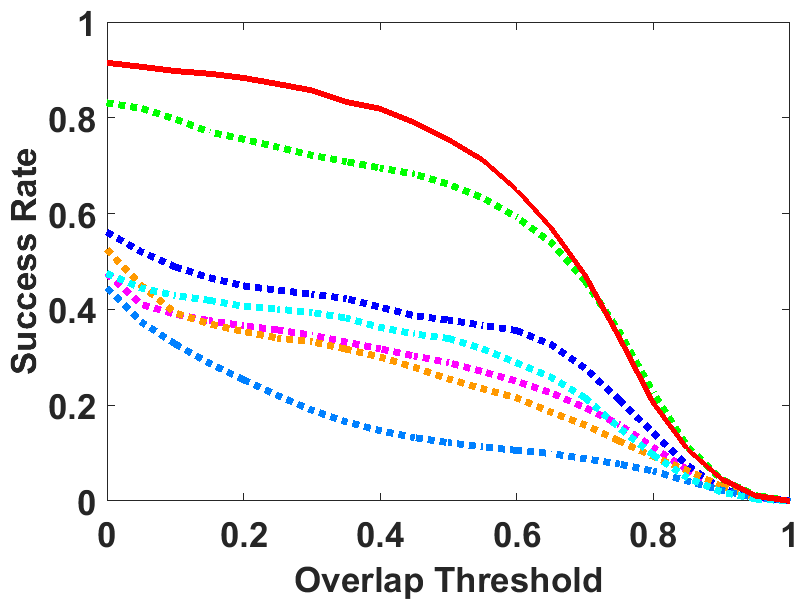}
  }\hfil
  \subfigure[Out-of-plane rotations]{
   \includegraphics[width=0.225\linewidth]{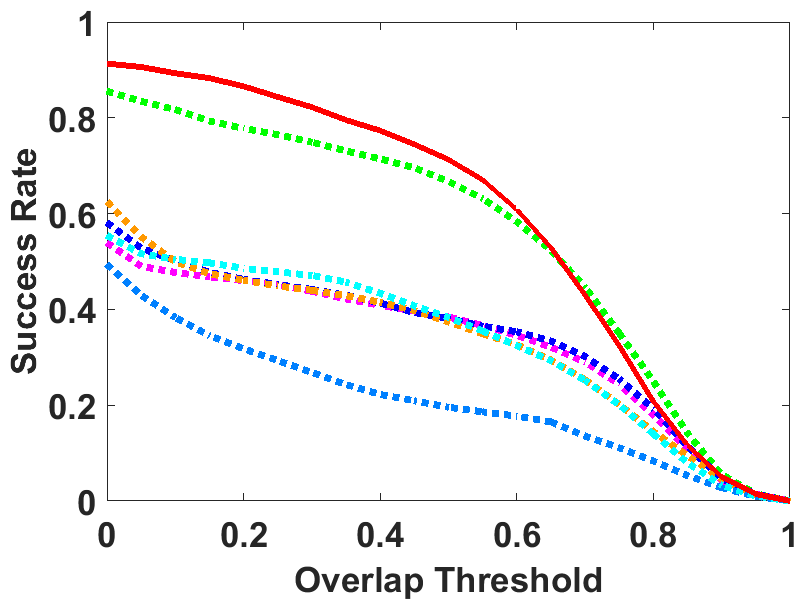}
  }\hfil
  \subfigure[In-plane rotations]{
   \includegraphics[width=0.225\linewidth]{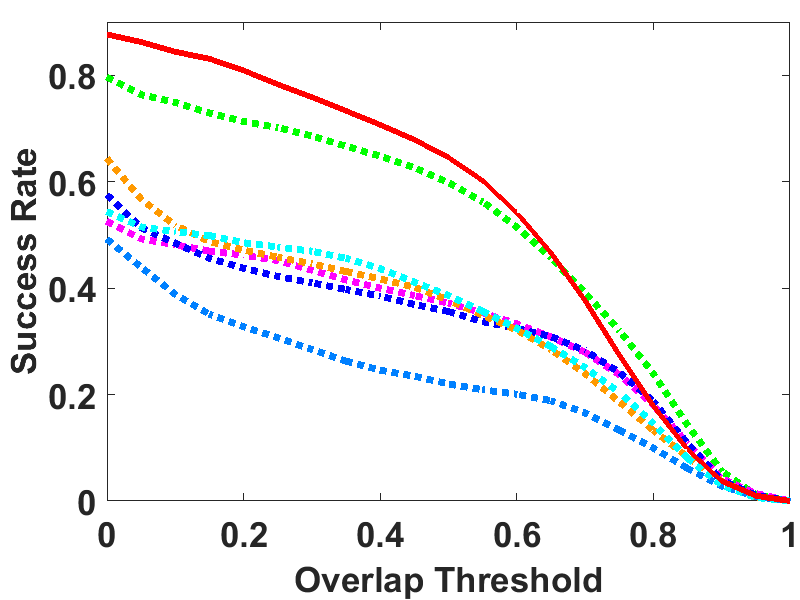}
  }\hfil
  \subfigure[Low resolutions]{
   \includegraphics[width=0.225\linewidth]{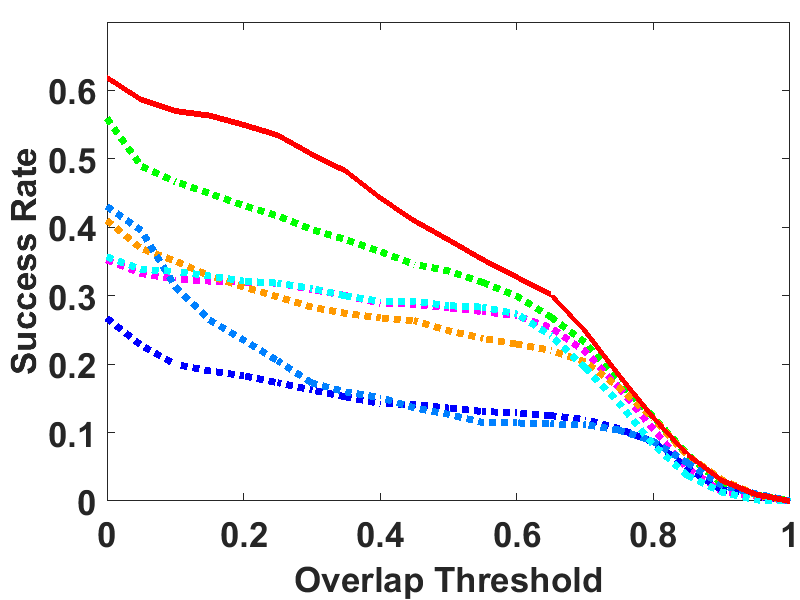}
  }\hfil
  \subfigure[Fast motion]{
   \includegraphics[width=0.225\linewidth]{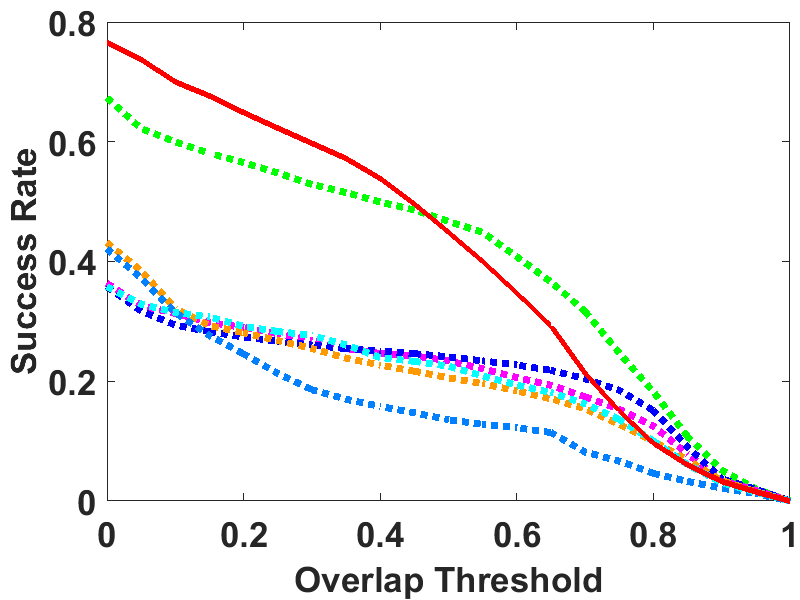}
  }\hfil
  \subfigure[Motion blur]{
   \includegraphics[width=0.225\linewidth]{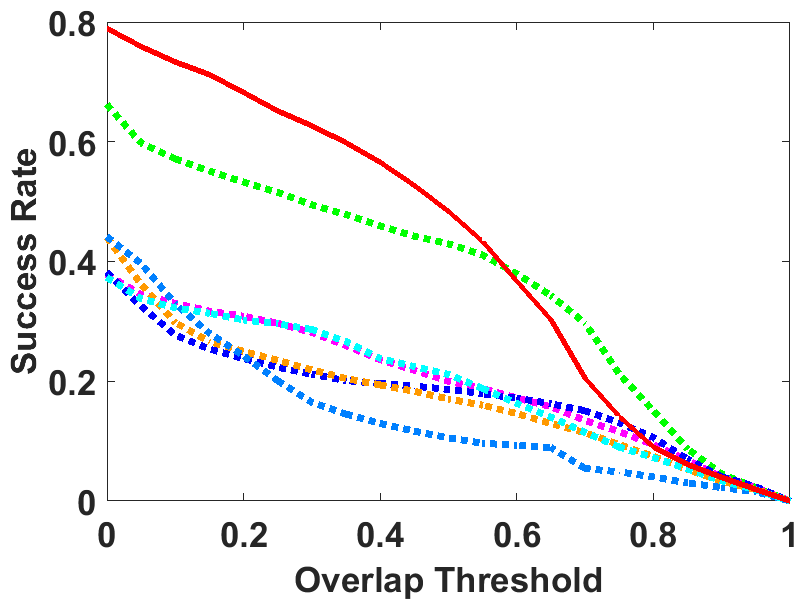}
  }\hfil
  \subfigure[Out of view]{
   \includegraphics[width=0.225\linewidth]{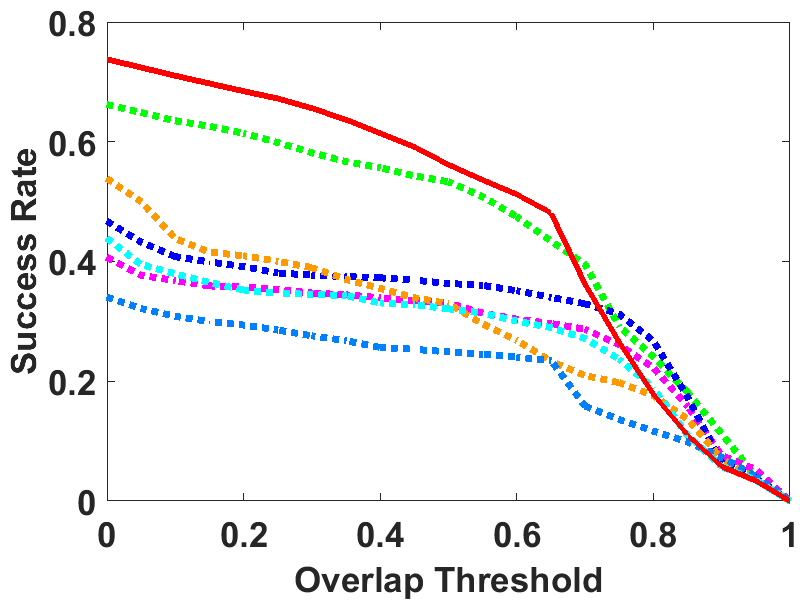}
  }\hfil
  \subfigure[Scale variations]{
   \includegraphics[width=0.225\linewidth]{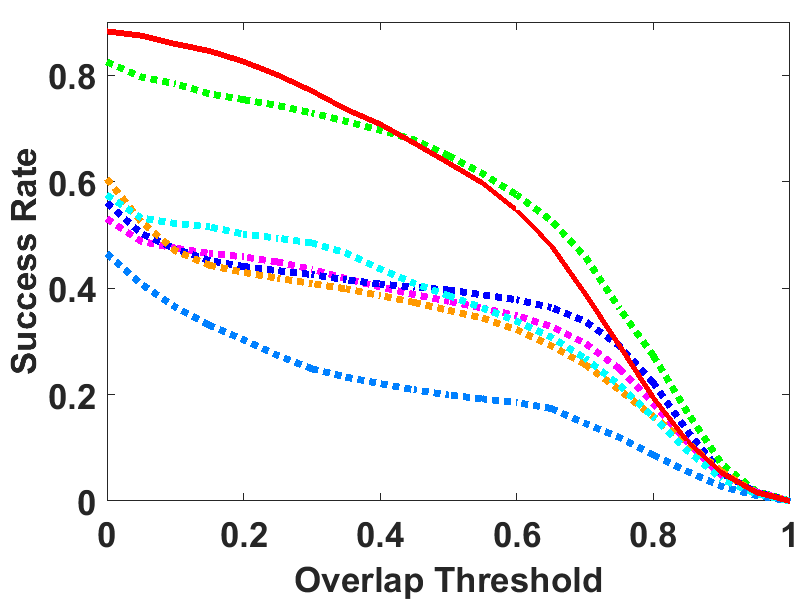}
  }\\
  \includegraphics[width=0.3\linewidth]{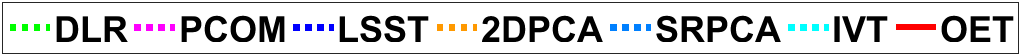}
\end{center}
  \caption{Tracking performance of the proposed tracker and other 6 state-of-the-art subspace trackers under various situations on OTB-100 benchmark.}
\label{fig:cases}
\end{figure*}
%

\section{Experiments}
The proposed tracker is implemented in MATLAB without any code optimizations on a PC with an Intel Xeon W3520 CPU at 2.67GHz. The ROIs of both the target and background samples are resized to $20\times20$ pixels and the histogram of oriented gradient features are extracted. The parameter $\lambda$ in \eqref{eq:E} is set to $10^{-4}$ in all the experiments. 50 target samples and 48 background samples are collected to form a training set and these samples are updated in each frame during tracking. The 50 target samples are collected from the 50 most recently localized targets as addressed above. The 48 background samples are collected from the ROIs with 2D shifting from the latest target region by $\{5,7,9,11,13,15\}$ pixels. 400 candidates are generated in each frame with the possibility corresponding to the target likelihood of the candidates in the last frame. The proposed tracker is named OET (Orthogonal Embedding Tracking) in this work, as shown in the experimental results.

\subsection{Benchmark Datasets and Baseline Trackers}
Four popular benchmark datasets are adopted to evaluate the proposed tracker thoroughly: OTB-50 \cite{Wu2013}, OTB-100 \cite{Wu2015}, MCT 2016 \cite{Sui2016mct}, and VOT 2016 \cite{vot2016}. The four benchmarks aim at different challenges and yield different evaluation protocols and criteria.

The OTB-50 and OTB-100 benchmarks include 50 and 100 fully labeled video sequences, respectively. The challenging situations are grouped into 11 categories like occlusion, illumination variation, and non-rigid deformation. One pass evaluation (OPE) is used as the main evaluation protocol on the two benchmarks. Precision and success rate curves are employed as the evaluation criteria. The precision at threshold 20 pixels and the area under curve (AUC) of the success rate plot are often reported as quantitative indexes for the tracking performance.

The MCT 2016 benchmark contains 20 challenging video sequences with a focus on people tracking. Similar to the OTB benchmarks, the MCT 2016 benchmark employs the OPE protocol. Tracking location errors and average overlap rate are adopted as the evaluation criteria. The former is consistent with the precision and indicates the accuracy of the tracking location center, whereas the latter measures the accuracy of the estimate for the target object scale.

The VOT 2016 benchmark consists of 60 short video sequences where the ground truth in each frame supports rotated bounding box. This benchmark allows tracking failures and thus adopts a resetting based evaluation protocol in addition to the OPE. A tracking failure is detected if the overlap between the tracking and ground truth bounding boxes drops to 0. Trackers under evaluation are re-initialized by the ground truth 5 frames after the failure point. The VOT 2016 benchmark uses overlap rate as the accuracy measurement and failure times as the robustness criterion. In addition, expected average overlap (EAO) is another important criterion for this benchmark, which provides an accuracy measurement from a statistical viewpoint.

We evaluate the performance of the proposed tracker and compare our results to extensive baseline trackers based on various tracking models, such as subspace learning, sparse learning, correlation filtering, and deep neural networks. In the experimental evaluations, we refer to the results of the baseline trackers, which are precomputed by their authors or obtained by running their authors' source codes.

\subsection{Evaluation Results}
\subsubsection{On OTB-50 Benchmark}
The proposed tracker is evaluated on the OTB-50 benchmark by comparing with 20 other state-of-the-art visual trackers. Fig. \ref{fig:result} shows the evaluation results where the 10 state-of-the-art trackers are based on correlation filtering (RCF \cite{Sui2016rcf}, DSST \cite{Danelljan2017tpami}, KCF \cite{Henriques2015}, and SAMF \cite{Li2014samf}), deep neural networks (HCFT \cite{Huang2015} and CNT \cite{Zhang2016c}), structured representation (SST \cite{Zhang2015a}), color attributes \cite{Danelljan2014} and complex model (TGPR \cite{Gao2014} and MEEM \cite{Zhang2014g}). The proposed tracker performs competitively against its 10 counterparts. It obtains the third best and the best results in terms of precision and success rate, respectively, and even performs better than the HCFT tracker, which relies on deep convolutional networks, on this benchmark in terms of success rate.

The proposed tracker is also evaluated on the OTB-50 benchmark by comparing with the up-to-date subspace trackers, including LLR \cite{Sui2016ijcv}, DSL \cite{Sui2015iccv}, SSL \cite{Sui2015tip}, DLR \cite{Sui2018ijcv}, \cite{Wang2014}, LSST \cite{Wang2013}, LRST \cite{Zhang2012b}, 2DPCA \cite{Wang2012}, SRPCA \cite{Wang2013a}, and IVT \cite{Ross2007}. Fig. \ref{fig:result} shows the tracking performance of the proposed tracker and the 8 latest subspace trackers. It is evident that the proposed tracker outperforms its 10 competing counterparts in terms of both precision and success rate. In particular, the proposed tracker yields 7\% increase in precision, and 6\% increase in AUC of success rate, in comparison with the second best tracker DLR. These evaluation results demonstrate the superior advantages of the proposed approach in subspace learning.

\begin{figure*}[t]
\begin{center}
  \includegraphics[width=0.35\linewidth]{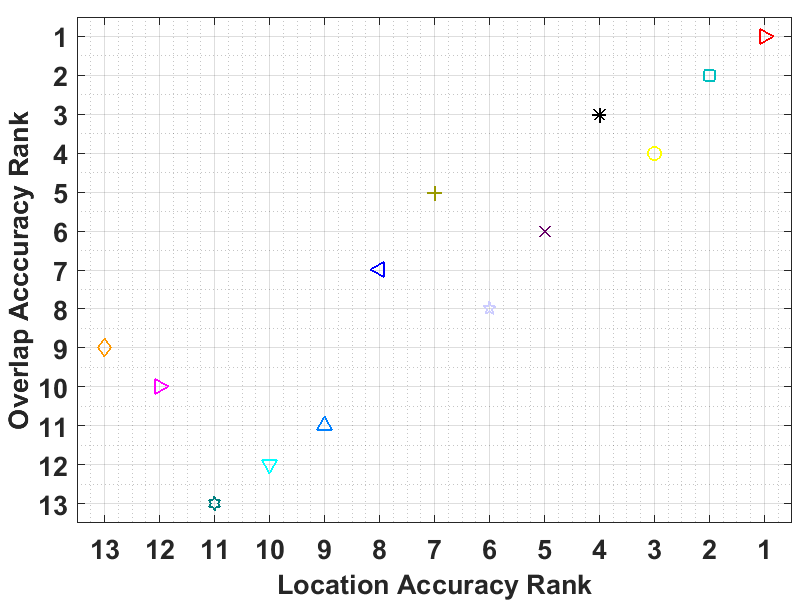}
 \hfil
 \includegraphics[width=0.35\linewidth]{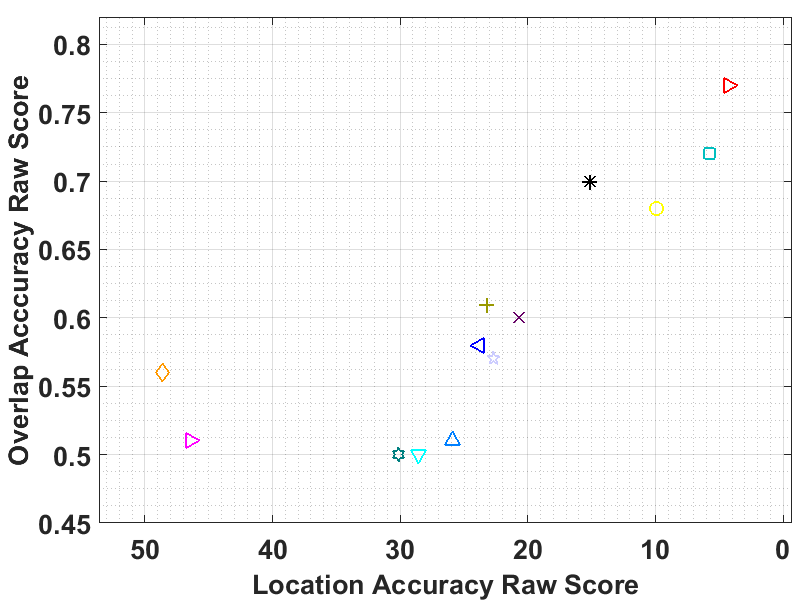}
 \hfil
 \includegraphics[width=0.224\linewidth]{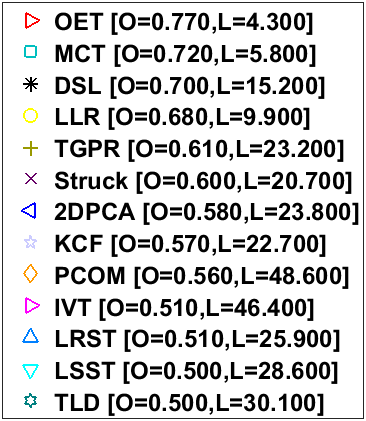}
\end{center}
 \caption{Tracking performance of the proposed tracker and its 12 competing counterparts on MCT 2016 benchmark. (left) Overlap-location rank plot; (middle) overlap-location raw score plot; (right) 'O' stands for average overlap and 'L' is the short for tracking location error in pixel.}
\label{fig:result20}
\end{figure*}
\begin{figure*}[t]
\begin{center}
   \includegraphics[width=0.475\linewidth]{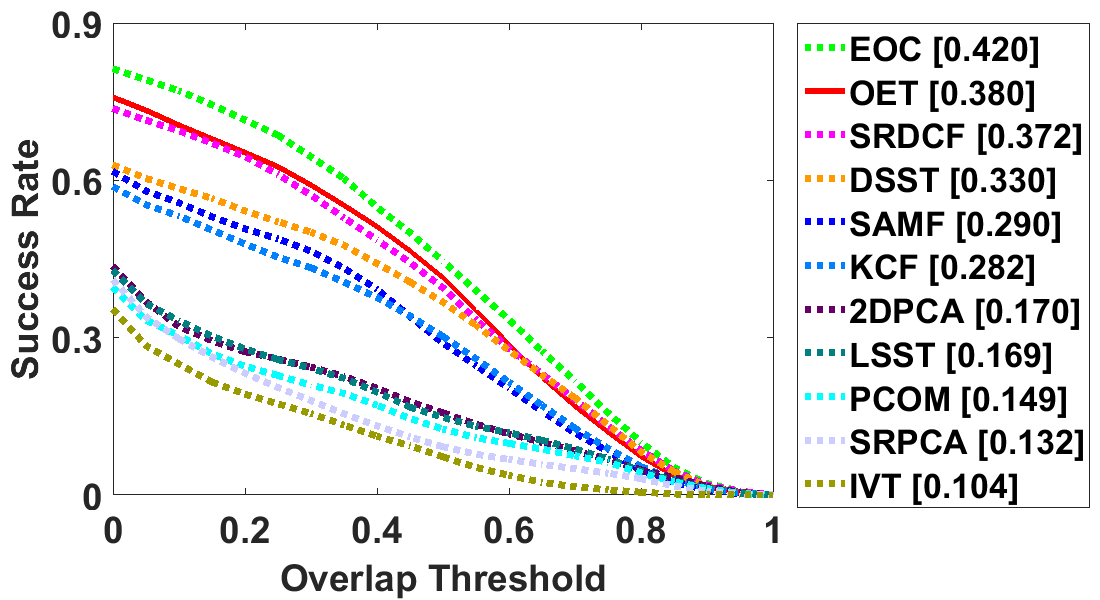}\hfil
   \includegraphics[width=0.35\linewidth]{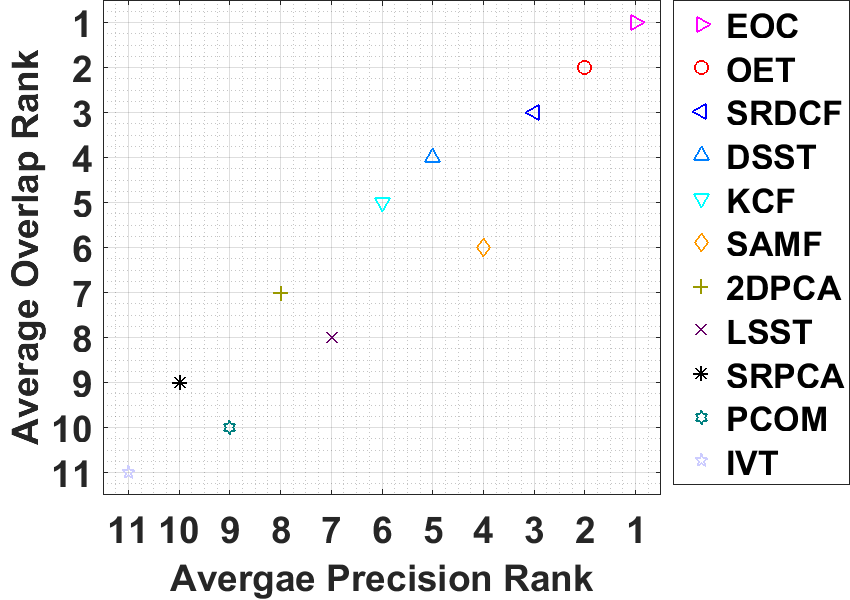}
\end{center}
  \caption{Tracking performance of the proposed tracker and other 10 state-of-the-art trackers on the VOT 2016 benchmark under one pass evaluation protocol. (left) success rate plot where the area under curve is reported in the legend; (right) overlap-precision rank plot.}
\label{fig:ope_vot2016}
\end{figure*}

\subsubsection{On OTB-100 Benchmark}
We evaluate the proposed tracker on the OTB-100 benchmark \cite{Wu2015}, as shown in Fig. \ref{fig:result_otb100}. Other 10 state-of-the-art trackers are referred to as the baselines, including 4 correlation filtering based trackers (DSST \cite{Danelljan2017tpami}, KCF \cite{Henriques2015}, SAMF \cite{Li2014samf}, CN \cite{Danelljan2014}, and CSK \cite{Henriques2012}) and 6 up-to-date subspace trackers (DLR \cite{Sui2018ijcv}, LSST \cite{Wang2013}, 2DPCA \cite{Wang2012}, PCOM \cite{Wang2014}, IVT \cite{Ross2007}, and SRPCA \cite{Wang2013a}). It can be seen that the proposed tracker outperforms all counterparts in terms of both the success rate and the precision, respectively. To be specific, the proposed tracker exceeds the DLR tracker, the best one among the 6 competing subspace trackers, by 5\% on this benchmark.

The performance of the proposed tracker is investigated under various challenging situations on the OTB-100 benchmark. 6 state-of-the-art subspace trackers are referred to as the baseline methods, including DLR \cite{Sui2018ijcv}, PCOM \cite{Wang2014}, LSST \cite{Wang2013}, 2DPCA \cite{Wang2012}, SRPCA \cite{Wang2013a}, and IVT \cite{Ross2007}. Fig. \ref{fig:cases} shows the evaluation results in all the 11 challenging situations, including occlusions, illumination variations, background clutters, non-rigid deformations, out-of-plane rotations, in-plane rotations, low resolutions, fast motion, motion blur, out of view, and scale variations. It is evident that the proposed tracker obtains competitive performance in these challenging situations. For example, the proposed tracker exhibits the best performance in the case of occlusions. This is attributed to: 1) the occlusions are formulated as outliers in the subspace learning, such that the subspace representations are robust against occlusions; and 2) the enhanced discrimination of the subspace representations improves the accuracy of the tracking in the presence of occlusions. In addition, the proposed tracker obtains competitive results in the case of scale variations. In the motion state variable used to describe the corresponding ROIs of the target and background samples in each frame, a component of scaling coefficient is introduced to adapt to the scale variations of the target appearance during tracking. The proposed tracker also outperforms its 6 competing counterparts in the case of non-rigid deformations. Subspace learning has been demonstrated to be effective to pose changes in visual tracking. The global deformations are well handled by the subspace learning, while the local deformations are considered as the noise corruption and can be absorbed by the additive sparse errors.

\subsubsection{On MCT 2016 Benchmark}
The proposed tracker is also evaluated on another popular benchmark, MCT 2016 benchmark \cite{Sui2016mct}, containing 20 fully labelled video sequences with various challenging situations. 12 state-of-the-art trackers are referred to as the baseline methods on this benchmark, including IVT \cite{Ross2007}, 2DPCA \cite{Wang2012}, PCOM \cite{Wang2014}, LRST \cite{Zhang2012b}, LSST \cite{Wang2013}, TLD \cite{Kalal2012}, Struck \cite{Hare2011}, TGPR \cite{Gao2014}, KCF \cite{Henriques2015}, DSL \cite{Sui2015iccv}, LLR \cite{Sui2016ijcv}, and MCT \cite{Sui2016mct}. Two accuracies, location accuracy and overlap accuracy, are considered in the evaluations on this benchmark, measured by tracking location error and average overlap rate, respectively.

Fig. \ref{fig:result20} shows the tracking performance of the proposed tracker and the 12 competing trackers. From the overlap-location rank plot, we can see that the proposed tracker outperforms its 12 competing counterparts. From the overlap-location raw score plot, it can be seen that the proposed tracker improves the overlap rate by 5\% on average compared with the second best tracker, the MCT tracker, on this benchmark.

\begin{figure*}[t]
\begin{center}
   \includegraphics[width=0.35\linewidth]{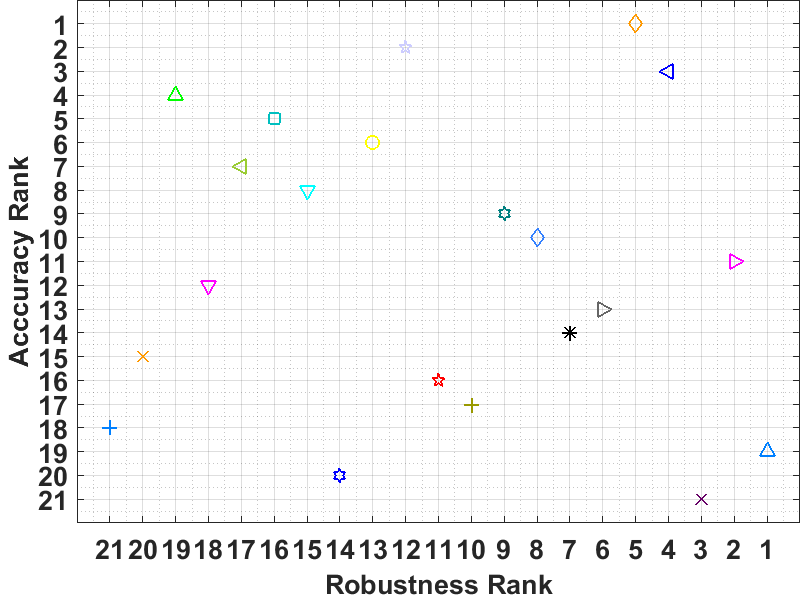}\hfil
   \includegraphics[width=0.35\linewidth]{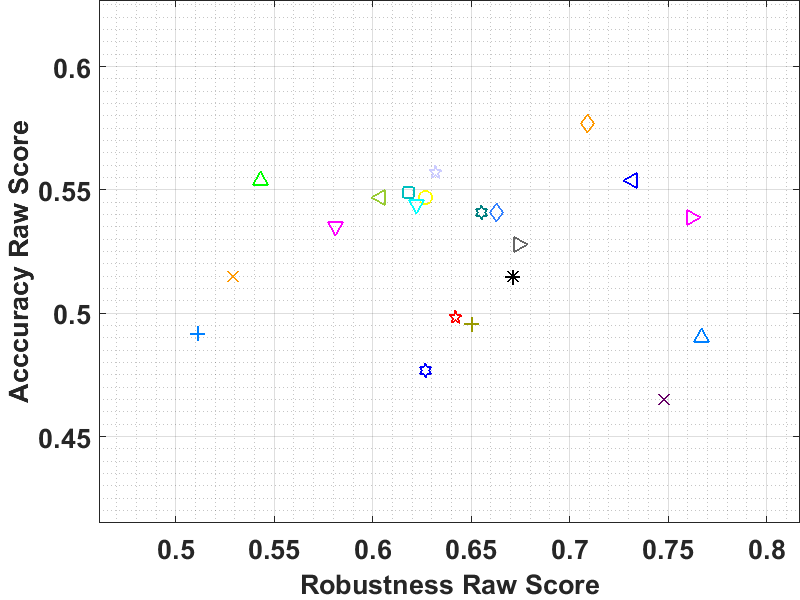}\hfil
   \includegraphics[width=0.162\linewidth]{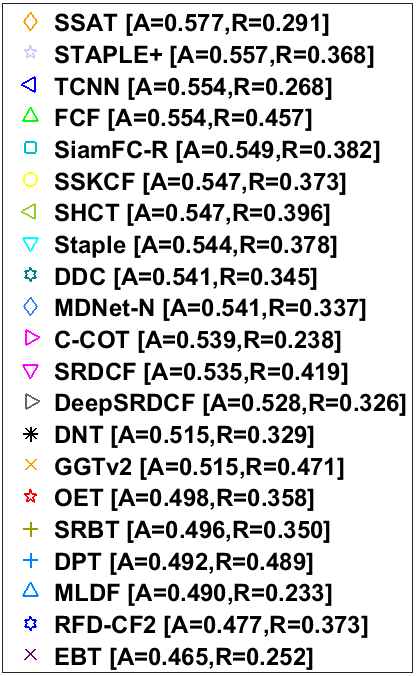}
\end{center}
  \caption{Tracking performance of the proposed tracker and the top trackers in VOT 2016 challenge in terms of accuracy and robustness on the VOT 2016 benchmark. (left) Accuracy-robustness rank plot; (middle) accuracy-robustness raw score plot; (right) 'A' stands for accuracy and 'R' is the short for robustness.}
\label{fig:ar_vot2016}
\end{figure*}
\begin{figure*}[t]
\begin{center}
   \includegraphics[width=0.35\linewidth]{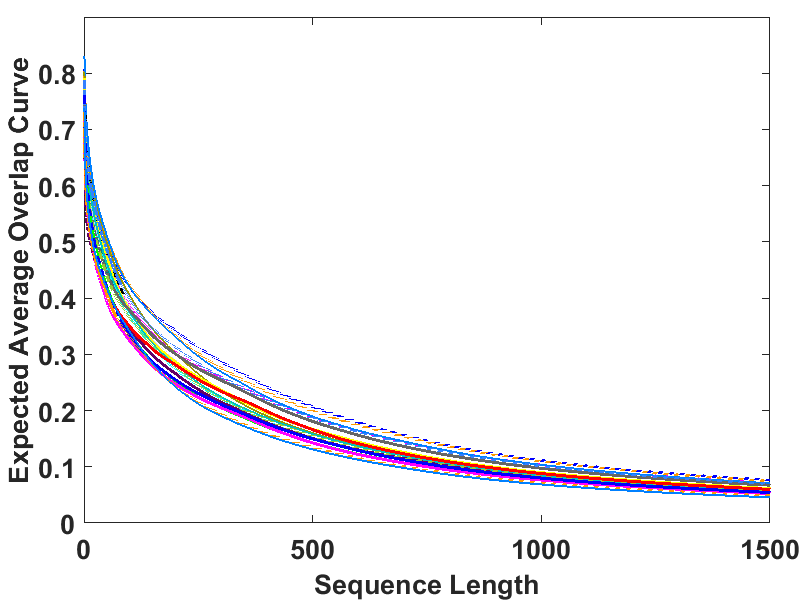}\hfil
   \includegraphics[width=0.35\linewidth]{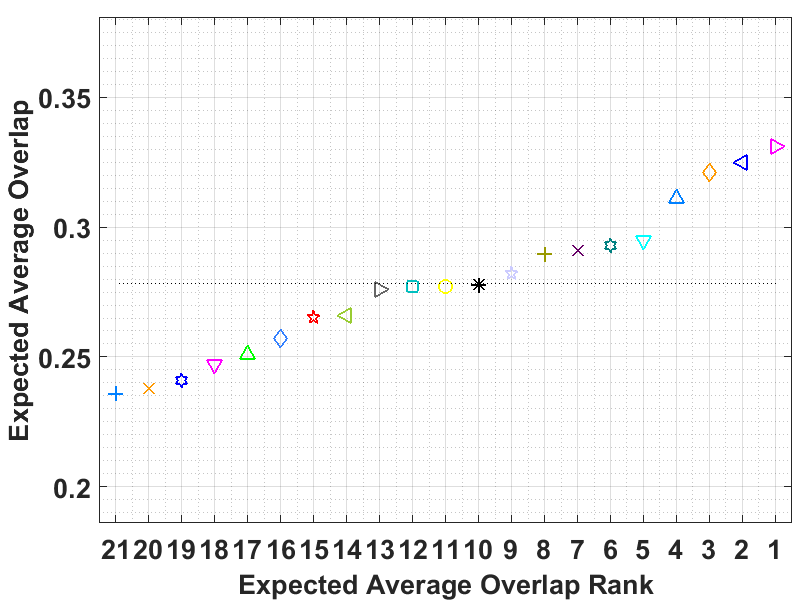}\hfil
   \includegraphics[width=0.108\linewidth]{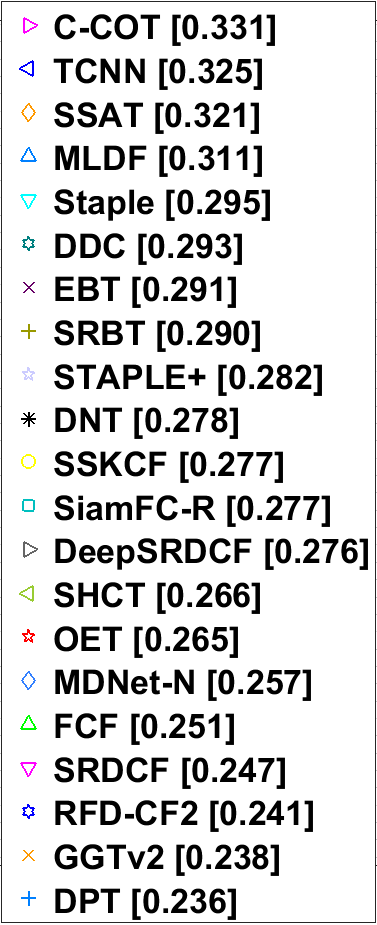}
\end{center}
  \caption{Tracking performance of the proposed tracker and the top 20 trackers in VOT 2016 challenge in terms of expected average overlap (EAO) on the VOT 2016 benchmark. (left) EAO curve plot; (middle) EAO rank plot where the dashed horizontal line indicates the average EAO score on all 21 trackers; (right) EAO scores are shown in the brackets.}
\label{fig:eao_vot2016}
\end{figure*}
\subsubsection{On VOT 2016 Benchmark}
The proposed tracker is also evaluated on the VOT 2016 benchmark \cite{vot2016}. We employ two evaluation protocols for this benchmark: one pass evaluation (OPE) and resetting based evaluation. In the OPE experiments, we refer to 10 baseline trackers, including 5 state-of-the-art trackers (EOC \cite{Danelljan2017cvpr}, SRDCF \cite{Danelljan2015iccv}, DSST \cite{Danelljan2017tpami}, KCF \cite{Henriques2015}, SAMF \cite{Li2014samf}), and 5 up-to-date subspace trackers (PCOM \cite{Wang2014}, LSST \cite{Wang2013}, 2DPAC \cite{Wang2012}, IVT \cite{Ross2007}, and SRPCA \cite{Wang2013a}). Fig. \ref{fig:ope_vot2016} shows the OPE results of the proposed tracker and its 10 competing counterparts. From the success rate plot, it can be seen that the proposed tracker yields the second best results, while it significantly (over 21\%) outperforms the 5 subspace trackers. From the the overlap-precision rank plot, we can see that the proposed tracker ranks second in terms of both the average overlap and the average precision, respectively.

The resetting based evaluation protocol is also used to evaluate the proposed tracker. A tracking failure is detected if the overlap rate in a frame drops to 0. The tracker is initialized by the ground truth bounding box after 5 frames when the tracking fails. Under this protocol, the accuracy is measured by the average overlap, and the numbers of tracking failure are mapped to real values within $\left[0,1\right]$, known as robustness. In addition to the above two criteria, expected average overlap (EAO) is another important metric for this protocol, which is considered from a statistical point of view for overlap rate.

In the evaluation experiments on this benchmark, we refer to the top 20 trackers in VOT 2016 challenge as the competing trackers. Fig. \ref{fig:ar_vot2016} shows the tracking performance of the proposed tracker and the 20 competing trackers in VOT 2016 challenge in terms of accuracy and robustness. From the accuracy-robustness rank plot, we can see that the proposed tracker ranks at 16 and 11 in terms of accuracy and robustness, respectively. It can be seen from the accuracy-robustness raw score plot that the proposed tracker performs competitively against the top 20 trackers with the accuracy score of 0.498 and the robustness score of 0.358, respectively. Fig. \ref{fig:eao_vot2016} shows the tracking performance of the proposed tracker and its 20 competing counterparts in terms of the EAO metric. It can be seen that the proposed tracker obtains good results that is comparable against the 20 competing trackers. Note that, although the proposed tracker ranks at 18, it achieves the average performance of the top 20 tracker, because the differences between the trackers ranking at from 12 to 18 are less than 0.1\%, \ie, the performance of these trackers is very close to each other.

\setlength{\tabcolsep}{3.5pt}
\begin{table}[t]
 \renewcommand{\arraystretch}{1.2}
 \caption{Performance gain of the proposed tracker over the best competing subspace trackers (shown in the last line) in AUC of success rate on different benchmarks.}
 \label{tab:gain}
 \centering
 \begin{tabular}{cccccc}
   \hline
   Benchmark & OTB-50 & OTB-100 & MCT 2016 & VOT 2016 & ~ \\
   \hline
   \# Videos & 50 & 100 & 20 & 60 & 230\\
   ~ & ~ & ~ & ~ & ~ & in total\\
   \hline
   Gain (\%) & 5.8 & 4.3 & 7.0 & 21.0 & 9.2\\
   ~ & DLR & DLR & DSL & 2DPCA & avg.\\
   \hline
 \end{tabular}
\end{table}
\subsubsection{In Summary}
We have conducted a large number of experiments on four benchmarks containing 230 video sequences in total to evaluate the proposed tracker. Overall, the proposed tracker performs competitively on the four benchmarks against the state-of-the-art trackers. Here we are particularly interested in the performance gain of the proposed tracker over the state-of-the-art subspace trackers. The proposed tracker significantly outperforms its peer subspace trackers on all the four benchmarks. We further investigate the performance difference between the proposed tracker and the best competing subspace tracker on each benchmark, which indicates the minimum performance increase of the proposed tracker over the state-of-the-art subspace trackers. Table \ref{tab:gain} shows the minimal performance gain on the four benchmarks in terms of AUC of success rate. Considering the numbers of the video sequences of the benchmarks, we compute the average performance gain on each video sequence. As shown in Table \ref{tab:gain}, the proposed tracker improves the tracking performance of subspace trackers more than 9\% on the 230 video sequences from the four benchmarks.

\subsection{Analysis of the Proposed Approach}
\begin{figure}[t]
\begin{center}
  \subfigure[]{
   \label{fig:embedding_a}
   \includegraphics[width=0.8\linewidth]{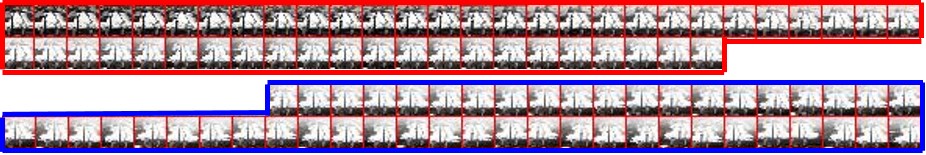}
  }
  \subfigure[]{
   \label{fig:embedding_b}
   \includegraphics[width=0.4\linewidth]{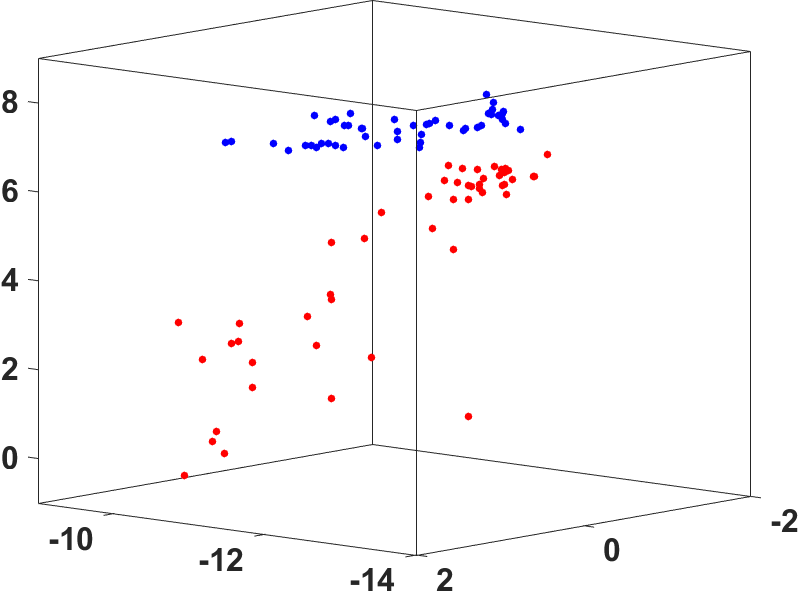}
  }
  \subfigure[]{
   \label{fig:embedding_c}
   \includegraphics[width=0.4\linewidth]{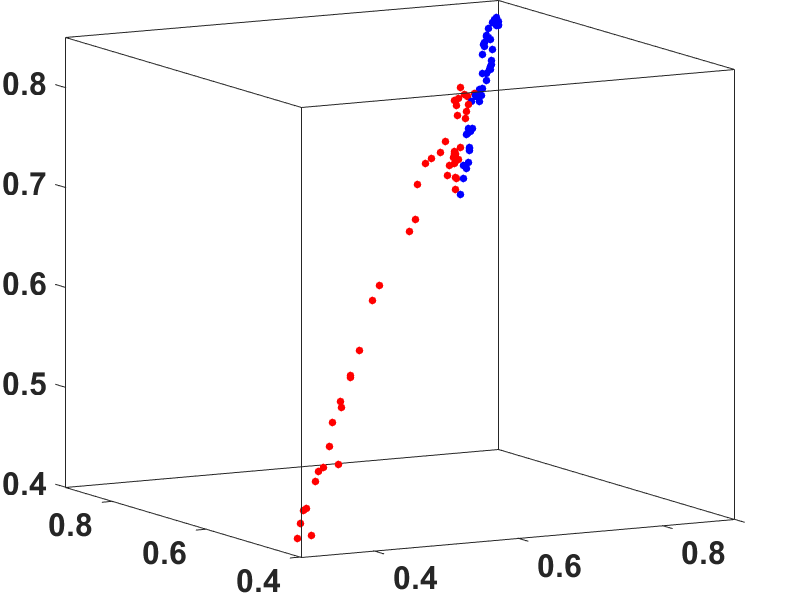}
  }
\end{center}
  \caption{(a) The target (marked in red) and background (in blue) samples in the case of illumination variation. Their subspace (b) representation (embedding) and (c) reconstruction clusters in the three-dimensional subspaces, respectively.}
\label{fig:embedding}
\end{figure}
\subsubsection{Reconstruction vs Embedding} First, we show a qualitative example in the case of illumination variation. The target and the background samples are shown in Fig. \ref{fig:embedding_a}. Their subspace embedding (used in the proposed approach) and reconstructions (used in \cite{Sui2015iccv}) are plotted in the 3D subspaces, as shown in Figs. \ref{fig:embedding_b} and \ref{fig:embedding_c}, respectively. It is evident that the subspace embedding of the two categories of samples are located in two clusters, leading to good linear separability in the low-dimensional subspace. In contrast, the subspace reconstructions belonging to two categories share almost the same direction of the principal axes, and they are hard to be separated from each other. Thus, the proposed approach constructs a subspace with better discrimination capability than DSL \cite{Sui2015iccv} in this example.

\begin{figure}[t]
\begin{center}
  \includegraphics[width=0.45\linewidth]{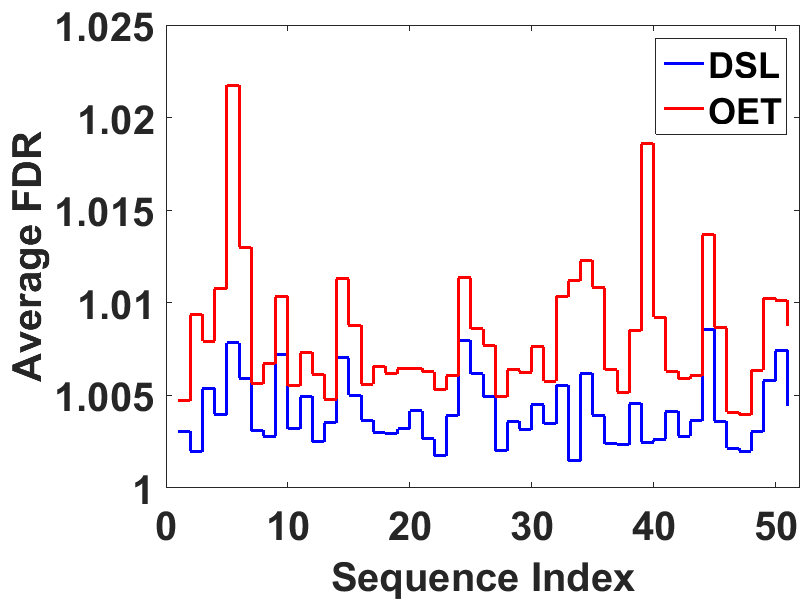}
  \hfil
  \includegraphics[width=0.45\linewidth]{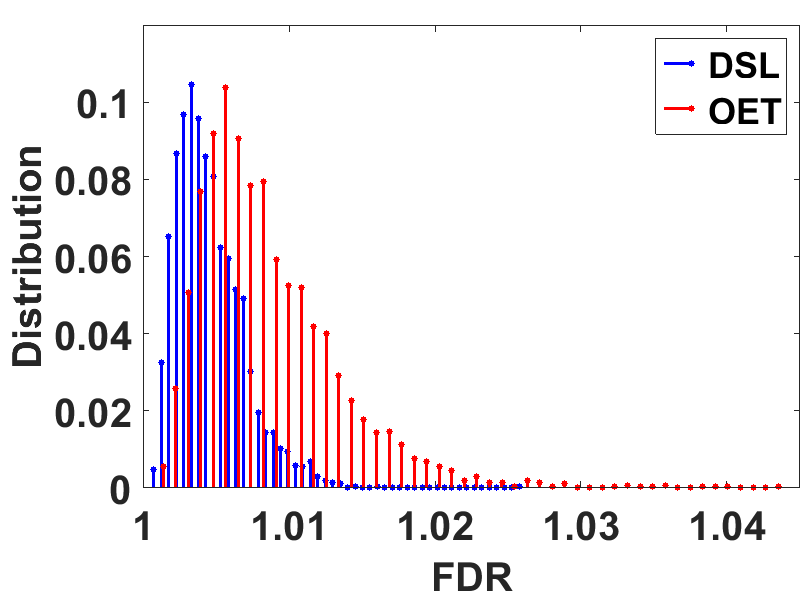}
\end{center}
  \caption{Discrimination capabilities of the DSL tracker \cite{Sui2015iccv} and the proposed tracker through (left) average FDR and (right) distributions of FDR on the OTB-50 benchmark.}
  \label{fig:fdr}
\end{figure}

Furthermore, Fig. \ref{fig:fdr} shows a quantitative demonstration on the OTB-50 benchmark. The fisher discriminative ratio (FDR) between the target and the background samples is used to measure the discrimination capability. It is evident that the proposed tracker obtains higher FDRs than \cite{Sui2015iccv} on this benchmark, which means better discrimination capability.

\subsubsection{Dimension Adaptivity vs Robustness Promotion} 
\begin{figure}[t]
  \centering
  \includegraphics[width=0.45\linewidth]{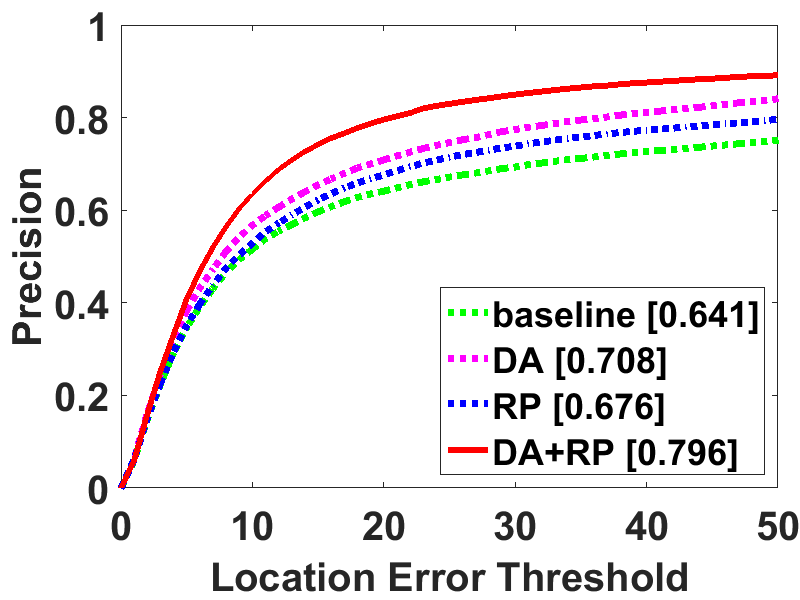}
  \hfil
  \includegraphics[width=0.45\linewidth]{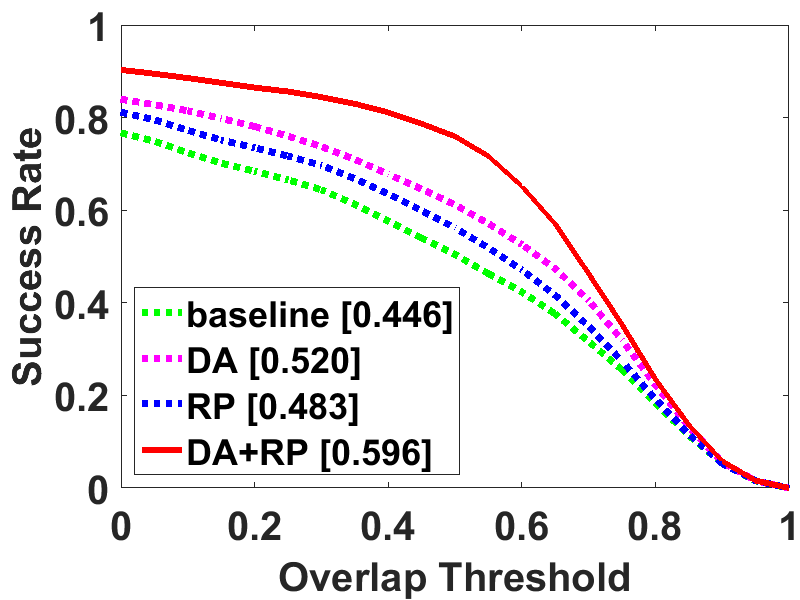}
  \caption{Performance of the dimension adaptivity (DA) and the robustness promotion (RP) on the OTB-50 benchmark.}
  \label{fig:da}
\end{figure}

We investigate the effectiveness of the dimension adaptivity (DA) and the robustness promotion (RP) on the OTB-50 benchmark, as shown in Fig. \ref{fig:da}. The baseline method adopts a subspace learning with a fixed dimension except RP. Because the dimension reduction obtained by our approach is 20$\%$ on average (as shown in Fig. \ref{fig:dim}), we fix the dimension at 20$\%$ of the original one. It is evident that the DA and the RP improve the tracking performance significantly.

\subsubsection{Subspace Embedding}
\begin{figure}[t]
\begin{center}
  \includegraphics[width=0.45\linewidth]{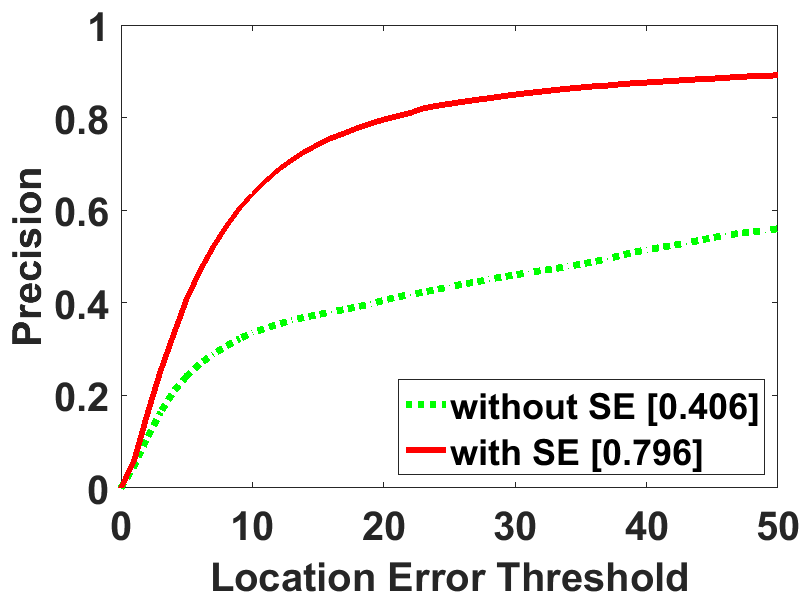}
  \hfil
  \includegraphics[width=0.45\linewidth]{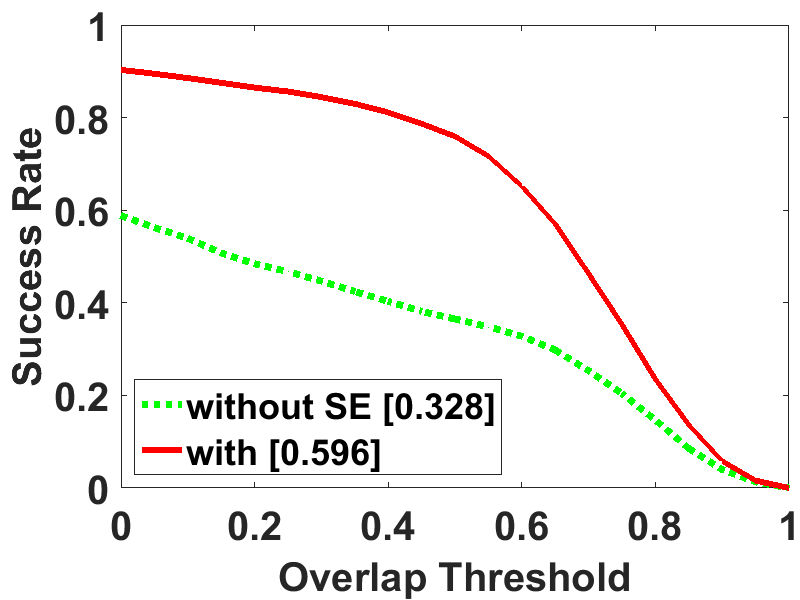}
\end{center}
  \caption{Tracking performance with and without subspace embedding on the OTB-50 benchmark.}
  \label{fig:se}
\end{figure}

We perform an ablation experiment to evaluate the performance with and without the subspace embedding on the OTB-50 benchmark, as shown in Fig. \ref{fig:se}. It can be seen that the tracking performance is significantly improved on this benchmark with the subspace embedding. The liner classifier is more effective and reliable over the subspace embedding, leading to better discrimination, which is a critical factor to visual tracking. Without the subspace embedding, the linear classifier is trained over the sample space that is higher-dimensional than where its embedding counterpart resides. Since the training sets are of the same size, the higher dimension puts much higher pressure on the training of the linear classifier, making the linear separability more difficult to reach.

\subsubsection{Dimension Reduction} In order to measure to what extent the dimension is reduced, a new criterion, \emph{the dimension reduction ratio} (DRR), is defined as $\frac{d}{min\left(m,n\right)}$, where the sample matrix is of size $m\times n$ and $d$ is the reduced dimension. Fig. \ref{fig:dim} shows the DRRs on each video sequence of the OTB-50 benchmark. It is evident that, on average, the dimension of the learned subspace is reduced to about 20\% of its original dimension. This result demonstrates that the proposed approach can effectively reduce the dimension in the subspace learning, making the subspace embedding more discriminative.

\subsubsection{Buffer Size in Online Update} 
\begin{figure}[t]
\begin{center}
  \subfigure[]{
   \label{fig:dim}
   \includegraphics[width=0.45\linewidth]{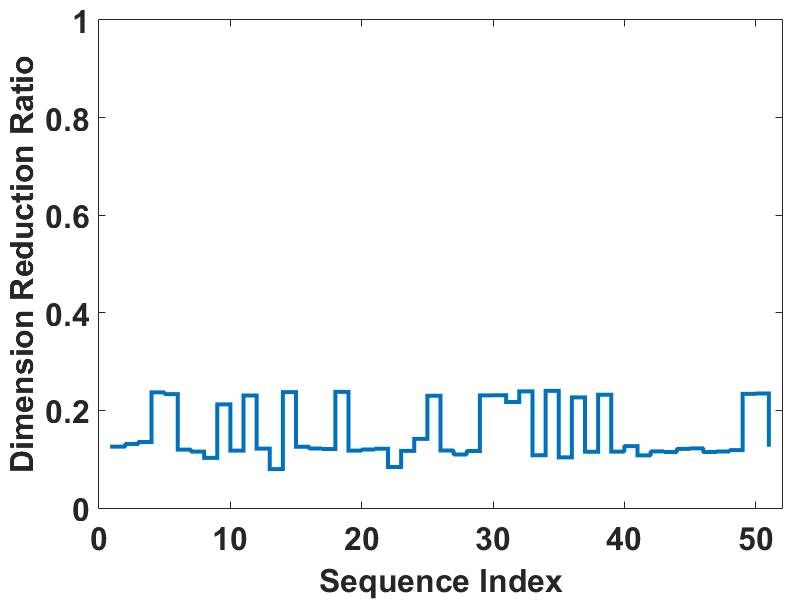}
  }
  \subfigure[]{
   \label{fig:ldr}
   \includegraphics[width=0.45\linewidth]{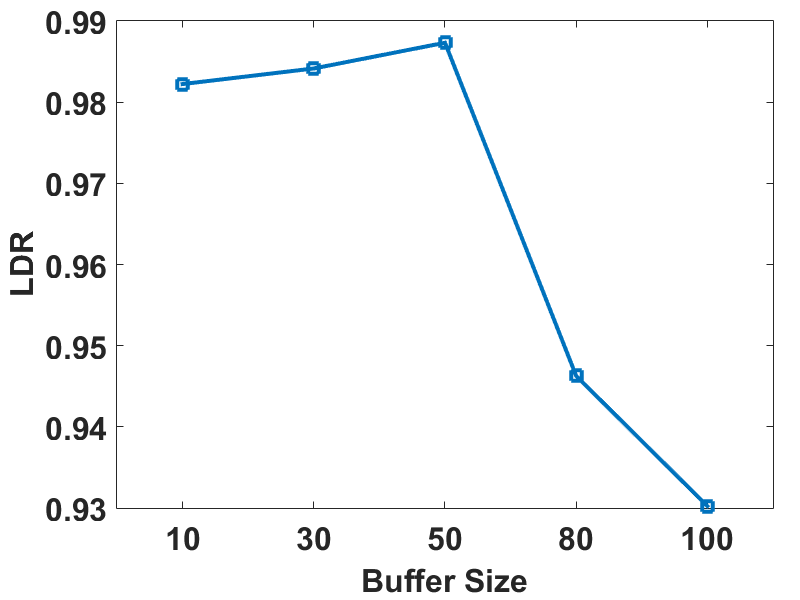}
  }
\end{center}
  \caption{Investigations on (a) dimension reduction ratio and (b) buffer size on the OTB-50 benchmark.}
\end{figure}

The buffer size is set to 50 for the online update. A smaller size yields lower dimension, whereas a larger size leads to better discrimination. We investigate the influence of the buffer size by defining a criterion, \emph{low-dimension discrimination ratio} (LDR) $d/l$, where $l$ is the low-dimension degree defined in \cite{Sui2016ijcv} and $d$ is the distance between the means of positive and negative samples. Larger LDRs indicate lower dimension and better discrimination. The influence of different buffer sizes is investigated on the OTB-50 benchmark, as shown in Fig. \ref{fig:ldr}, where we can see that the best performance is achieved when setting the buffer size at 50.

\section{Discussions}
\subsection{Computational Complexity}
The proposed tracker is implemented in MATLAB without any code optimizations. Its average running speed is 2 frames per second. The computational burden mainly comes from the SVD and matrix inversion for the likelihood evaluation of each target candidate in target localization, yielding a cubic order complexity $\mathcal{O}\left(n^3\right)$. Note that because the computational cost of the target localization is linearly proportional to the number of the target candidates, utilizing the independence of the candidates, the proposed tracker can be significantly sped up (theoretically by the number of candidates) by paralleling the likelihood computations of all the 400 candidates in the target localization.

\subsection{Limitations}
The proposed subspace embedding learning approach is driven by a Bayesian inference frame for the tracking. In the motion model, we assume the 2D translations to be Gaussian with a zero mean and a variance of 4 pixels. Because 95\% samples drawn from the Gaussian are located within the range of 3 times standard deviation, the motion model can cover less than 6 pixels of the 2D translations. If the motion between two consecutive frames is larger than 6 pixels, the proposed tracker might fail since the candidates are rarely generated in the regions far away from the target location. It indicates that the proposed tracker might perform poorly in the case of fast motion. To alleviate this limitation, we can set a large value to the variance of the 2D translation to enlarge the search area for the target candidates. However, much more candidates are correspondingly required to ensure high tracking accuracy over the large search region. As analyzed above, the computational cost increases linearly with the number of the candidates. To make a trade-off between the tracking accuracy and computational efficiency, we use 400 candidates and set the variance to 4 pixels for the 2D translation motion model. With this configuration, the proposed tracker perform well in the case of fast motion on most benchmarks.

\section{Conclusion}
In this work, motivated by the assumption that the recently localized targets and their immediately surrounding background reside in a low-dimensional subspace, a dimension-adaptive subspace with good discrimination capability has been constructed by injecting a discriminative approach that reliably separates the target from the background into the subspace learning by means of joint learning. Subspace embedding has been leveraged to make dimension reduction for the discrimination learning. A new tracking model has been constructed and a new formulation has been proposed to solve the subspace embedding by leveraging a HSIC method. Extensive experimental results on four standard tracking benchmarks have demonstrated that the proposed tracker performs competitively against the state-of-the-art trackers, and significantly improves the performance of up-to-date subspace trackers.
\newpage

%



\ifCLASSOPTIONcompsoc
\else
  \section*{Acknowledgment}
\fi


\ifCLASSOPTIONcaptionsoff
  \newpage
\fi



\bibliographystyle{IEEEtran}
\bibliography{refs}

\begin{thebibliography}{10}
\providecommand{\url}[1]{#1}
\csname url@samestyle\endcsname
\providecommand{\newblock}{\relax}
\providecommand{\bibinfo}[2]{#2}
\providecommand{\BIBentrySTDinterwordspacing}{\spaceskip=0pt\relax}
\providecommand{\BIBentryALTinterwordstretchfactor}{4}
\providecommand{\BIBentryALTinterwordspacing}{\spaceskip=\fontdimen2\font plus
\BIBentryALTinterwordstretchfactor\fontdimen3\font minus
  \fontdimen4\font\relax}
\providecommand{\BIBforeignlanguage}[2]{{%
\expandafter\ifx\csname l@#1\endcsname\relax
\typeout{** WARNING: IEEEtran.bst: No hyphenation pattern has been}%
\typeout{** loaded for the language `#1'. Using the pattern for}%
\typeout{** the default language instead.}%
\else
\language=\csname l@#1\endcsname
\fi
#2}}
\providecommand{\BIBdecl}{\relax}
\BIBdecl

\bibitem{Yilmaz2006}
A.~Yilmaz, O.~Javed, and M.~Shah, ``{Object tracking: A Survey},'' \emph{ACM
  Computing Surveys}, vol.~38, no.~4, pp. 13--57, 2006.

\bibitem{Smeulders2014}
A.~W.~M. Smeulders, D.~M. Chu, R.~Cucchiara, S.~Calderara, A.~Dehghan, and
  M.~Shah, ``{Visual Tracking: An Experimental Survey},'' \emph{IEEE TPAMI},
  vol.~36, no.~7, pp. 1442--1468, 2014.

\bibitem{sui2016real}
Y.~Sui, Z.~Zhang, G.~Wang, Y.~Tang, and L.~Zhang, ``Real-time visual tracking:
  Promoting the robustness of correlation filter learning,'' in \emph{European
  conference on computer vision}.\hskip 1em plus 0.5em minus 0.4em\relax
  Springer, 2016, pp. 662--678.

\bibitem{bharati2018real}
S.~P. Bharati, Y.~Wu, Y.~Sui, C.~Padgett, and G.~Wang, ``Real-time obstacle
  detection and tracking for sense-and-avoid mechanism in uavs,'' \emph{IEEE
  Transactions on Intelligent Vehicles}, vol.~3, no.~2, pp. 185--197, 2018.

\bibitem{wu2017vision}
Y.~Wu, Y.~Sui, and G.~Wang, ``Vision-based real-time aerial object localization
  and tracking for uav sensing system,'' \emph{IEEE Access}, vol.~5, pp.
  23\,969--23\,978, 2017.

\bibitem{Ross2007}
D.~A. Ross, J.~Lim, R.-S. Lin, and M.-H. Yang, ``{Incremental Learning for
  Robust Visual Tracking},'' \emph{International Journal of Computer Vision
  (IJCV)}, vol.~77, no. 1-3, pp. 125--141, 2007.

\bibitem{Li2007}
X.~Li, W.~Hu, and Z.~Zhang, ``{Robust visual tracking based on incremental
  tensor subspace learning},'' in \emph{ICCV}, 2007.

\bibitem{Kwon2010}
J.~Kwon and K.~Lee, ``{Visual tracking decomposition},'' in \emph{CVPR}, 2010.

\bibitem{Wang2012}
D.~Wang and H.~Lu, ``{Object tracking via 2DPCA and L1-regularization},''
  \emph{IEEE Signal Processing Letters}, vol.~19, no.~11, pp. 711--714, 2012.

\bibitem{Wang2013}
D.~Wang, H.~Lu, and M.-H. Yang, ``{Least Soft-thresold Squares Tracking},'' in
  \emph{CVPR}, 2013.

\bibitem{Wang2013a}
------, ``{Online object tracking with sparse prototypes},'' \emph{IEEE
  Transactions on Image Processing (TIP)}, vol.~22, no.~1, pp. 314--325, 2013.

\bibitem{Wang2014}
D.~Wang and H.~Lu, ``{Visual Tracking via Probability Continuous Outlier
  Model},'' in \emph{CVPR}, 2014.

\bibitem{Sui2015tip}
Y.~Sui, S.~Zhang, and L.~Zhang, ``{Robust Visual Tracking via Sparsity-Induced
  Subspace Learning},'' \emph{IEEE Transactions on Image Processing (TIP)},
  vol.~24, no.~12, pp. 4686--4700, 2015.

\bibitem{Sui2015iccv}
Y.~Sui, Y.~Tang, and L.~Zhang, ``{Discriminative Low-Rank Tracking},'' in
  \emph{ICCV}, 2015.

\bibitem{Sui2016ijcv}
Y.~Sui and L.~Zhang, ``{Robust Tracking via Locally Structured
  Representation},'' \emph{International Journal of Computer Vision (IJCV)},
  vol. 119, no.~2, pp. 110--144, 2016.

\bibitem{sui2019sparse}
Y.~Sui, G.~Wang, and L.~Zhang, ``Sparse subspace clustering via low-rank
  structure propagation,'' \emph{Pattern Recognition}, vol.~95, pp. 261--271,
  2019.

\bibitem{Hager1996}
G.~D. Hager and P.~N. Belhumeur, ``{Real-Time Tracking of Image Regions with
  Changes in Geometry and Illumination},'' in \emph{CVPR}, 1996.

\bibitem{Kriegmant1996}
P.~N. Belhumeur and D.~J. Kriegmant, ``{What is the Set of Images of an Object
  Under All Possible Lighting Conditions?}'' in \emph{CVPR}, 1996.

\bibitem{Duda2001}
R.~Duda, P.~Hart, and D.~Stork, \emph{{Pattern Classification, 2nd
  Edition}}.\hskip 1em plus 0.5em minus 0.4em\relax Wiley, 2001.

\bibitem{Zhang2012a}
T.~Zhang, B.~Ghanem, and S.~Liu, ``{Robust Visual Tracking via Multi-Task
  Sparse Learning},'' in \emph{CVPR}, 2012.

\bibitem{Zhang2012b}
T.~Zhang, B.~Ghanem, S.~Liu, and N.~Ahuja, ``{Low-rank sparse learning for
  robust visual tracking},'' in \emph{ECCV}, 2012.

\bibitem{Candes2011}
E.~J. Cand{\`{e}}s, X.~Li, Y.~Ma, and J.~Wright, ``{Robust principal component
  analysis?}'' \emph{Journal of the ACM}, vol.~58, no.~3, pp. 1--37, 2011.

\bibitem{Sui2018ijcv}
Y.~Sui, Y.~Tang, L.~Zhang, and G.~Wang, ``{Visual Tracking via Subspace
  Learning: A Discriminative Approach},'' \emph{International Journal of
  Computer Vision (IJCV)}, vol. 126, no.~5, pp. 515--536, 2018.

\bibitem{Comaniciu2003}
D.~Comaniciu, S.~Member, and V.~Ramesh, ``{Kernel-Based Object Tracking},''
  \emph{IEEE TPAMI}, vol.~25, no.~5, pp. 564--577, 2003.

\bibitem{Mei2009}
X.~Mei and H.~Ling, ``{Robust visual tracking using L1 minimization},'' in
  \emph{ICCV}, 2009.

\bibitem{Zhang2016}
T.~Zhang, A.~Bibi, and B.~Ghanem, ``{In Defense of Sparse Tracking : Circulant
  Sparse Tracker},'' in \emph{CVPR}, 2016.

\bibitem{Wang2013b}
N.~Wang, J.~Wang, and D.~Yeung, ``{Online Robust Non-negative Dictionary
  Learning for Visual Tracking},'' in \emph{ICCV}, 2013.

\bibitem{Sui2015pr}
Y.~Sui, X.~Zhao, S.~Zhang, X.~Yu, S.~Zhao, and L.~Zhang, ``{Self-expressive
  tracking},'' \emph{Pattern Recognition (PR)}, vol.~48, no.~9, pp. 2872--2884,
  2015.

\bibitem{Sui2015spl}
Y.~Sui and L.~Zhang, ``{Visual Tracking via Locally Structured Gaussian Process
  Regression},'' \emph{IEEE Signal Processing Letters}, vol.~22, no.~9, pp.
  1331--1335, 2015.

\bibitem{Kalal2012}
Z.~Kalal, K.~Mikolajczyk, and J.~Matas, ``{Tracking-learning-detection},''
  \emph{IEEE TPAMI}, vol.~34, no.~7, pp. 1409--1422, 2012.

\bibitem{Babenko2011}
B.~Babenko, S.~Member, M.-H. Yang, and S.~Member, ``{Robust Object Tracking
  with Online Multiple Instance Learning},'' \emph{IEEE TPAMI}, vol.~33, no.~8,
  pp. 1619--1632, 2011.

\bibitem{sui2018joint}
Y.~Sui, G.~Wang, and L.~Zhang, ``Joint correlation filtering for visual
  tracking,'' \emph{IEEE Transactions on Circuits and Systems for Video
  Technology}, vol.~30, no.~1, pp. 167--178, 2018.

\bibitem{Sui2016rcf}
Y.~Sui, Z.~Zhang, G.~Wang, Y.~Tang, and L.~Zhang, ``{Real-Time Visual Tracking:
  Promoting the Robustness of Correlation Filter Learning},'' in \emph{ECCV},
  2016.

\bibitem{Mueller2017}
M.~Mueller, N.~Smith, and B.~Ghanem, ``{Context-Aware Correlation Filter
  Tracking},'' in \emph{CVPR}, 2017.

\bibitem{Sui2018tcyb}
Y.~Sui, G.~Wang, and L.~Zhang, ``{Correlation Filter Learning towards Peak
  Strength for Visual Tracking},'' \emph{IEEE Transactions on Cybernetics
  (TCyb)}, vol.~48, no.~4, pp. 1290--1303, 2018.

\bibitem{bharati2016fast}
S.~P. Bharati, S.~Nandi, Y.~Wu, Y.~Sui, and G.~Wang, ``Fast and robust object
  tracking with adaptive detection,'' in \emph{2016 IEEE 28th international
  conference on tools with artificial intelligence (ICTAI)}.\hskip 1em plus
  0.5em minus 0.4em\relax IEEE, 2016, pp. 706--713.

\bibitem{Huang2015}
C.~Ma, J.-B. Huang, X.~Yang, and M.-H. Yang, ``{Hierarchical Convolutional
  Features for Visual Tracking},'' in \emph{ICCV}, 2015.

\bibitem{Qi2016}
Y.~Qi, S.~Zhang, L.~Qin, H.~Yao, Q.~Huang, J.~Lim, and M.-H. Yang, ``{Hedged
  Deep Tracking},'' in \emph{CVPR}, 2016.

\bibitem{Nam2016}
H.~Nam and B.~Han, ``{Learning Multi-Domain Convolutional Neural Networks for
  Visual Tracking},'' in \emph{CVPR}, 2016.

\bibitem{Choi2017}
J.~Choi, H.~J. Chang, S.~Yun, T.~Fischer, Y.~Demiris, and J.~Y. Choi,
  ``{Attentional Correlation Filter Network for Adaptive Visual Tracking},'' in
  \emph{CVPR}, 2017.

\bibitem{Han2017}
B.~Han, J.~Sim, and H.~Adam, ``{BranchOut: Regularization for Online Ensemble
  Tracking with Convolutional Neural Networks},'' in \emph{CVPR}, 2017.

\bibitem{zhang2020real}
X.~Zhang, T.~Zhang, Y.~Yang, Z.~Wang, and G.~Wang, ``Real-time golf ball
  detection and tracking based on convolutional neural networks,'' in
  \emph{2020 IEEE International Conference on Systems, Man, and Cybernetics
  (SMC)}.\hskip 1em plus 0.5em minus 0.4em\relax IEEE, 2020, pp. 2808--2813.

\bibitem{Zhong2012}
W.~Zhong, H.~Lu, and M.-H. Yang, ``{Robust Object Tracking via Sparsity-based
  Collaborative Model},'' in \emph{CVPR}, 2012.

\bibitem{Sui2018tip}
Y.~Sui, G.~Wang, L.~Zhang, and M.-H. Yang, ``{Exploiting Spatial-Temporal
  Locality of Tracking via Structured Dictionary Learning},'' \emph{IEEE
  Transactions on Image Processing (TIP)}, vol.~27, no.~3, pp. 1282--1296,
  2018.

\bibitem{Barshan2011}
E.~Barshan, A.~Ghodsi, Z.~Azimifar, and M.~Z. Jahromi, ``{Supervised principal
  component analysis: Visualization, classification and regression on subspaces
  and submanifolds},'' \emph{Pattern Recognition (PR)}, vol.~44, no.~7, pp.
  1357--1371, 2011.

\bibitem{Cai2010}
J.~Cai, E.~Cand{\`{e}}s, and Z.~Shen, ``{A singular value thresholding
  algorithm for matrix completion},'' \emph{SIAM Journal on Optimization},
  vol.~20, no.~4, pp. 1956--1982, 2010.

\bibitem{Beck2009}
A.~Beck and M.~Teboulle, ``{A Fast Iterative Shrinkage-Thresholding Algorithm
  for Linear Inverse Problems},'' \emph{SIAM Journal on Imaging Sciences},
  vol.~2, no.~1, pp. 183--202, 2009.

\bibitem{Wu2013}
Y.~Wu, J.~Lim, and M.-H. Yang, ``{Online Object Tracking: A Benchmark},'' in
  \emph{CVPR}, 2013.

\bibitem{Wu2015}
Y.~Wu, J.~Lim, and M.~H. Yang, ``{Object tracking benchmark},'' \emph{IEEE
  TPAMI}, vol.~37, no.~9, pp. 1834--1848, 2015.

\bibitem{Sui2016mct}
Y.~Sui, G.~Wang, Y.~Tang, and L.~Zhang, ``{Tracking Completion},'' in
  \emph{ECCV}, 2016.

\bibitem{vot2016}
M.~Kristan, A.~Leonardis, J.~Matas, M.~Felsberg, R.~Pflugfelder, L.~Cehovin,
  T.~Vojir, G.~Hager, A.~Lukezic, and G.~Fernandez, ``{The Visual Object
  Tracking VOT2016 challenge results},'' in \emph{ECCV Workshop}, 2016.

\bibitem{Danelljan2017tpami}
M.~Danelljan, G.~Hager, F.~S. Khan, and M.~Felsberg, ``{Discriminative Scale
  Space Tracking},'' \emph{IEEE TPAMI}, vol.~39, no.~8, pp. 1561--1575, 2017.

\bibitem{Henriques2015}
J.~Henriques, R.~Caseiro, P.~Martins, and J.~Batista, ``{High-Speed Tracking
  with Kernelized Correlation Filters},'' \emph{IEEE TPAMI}, vol.~37, no.~3,
  pp. 583--596, 2015.

\bibitem{Li2014samf}
Y.~Li and J.~Zhu, ``{A Scale Adaptive Kernel Correlation Filter Tracker with
  Feature Integration},'' in \emph{ECCV Workshop}, 2014.

\bibitem{Zhang2016c}
K.~Zhang, Q.~Liu, Y.~Wu, and M.-H. Yang, ``{Robust Visual Tracking via
  Convolutional Networks without Training},'' \emph{IEEE Transactions on Image
  Processing (TIP)}, vol.~25, no.~4, pp. 1779--1792, 2016.

\bibitem{Zhang2015a}
T.~Zhang, S.~Liu, C.~Xu, S.~Yan, B.~Ghanem, N.~Ahuja, and M.-H. Yang,
  ``{Structural Sparse Tracking},'' in \emph{CVPR}, 2015.

\bibitem{Danelljan2014}
M.~Danelljan, F.~S. Khan, M.~Felsberg, and J.~V.~D. Weijer, ``{Adaptive Color
  Attributes for Real-Time Visual Tracking},'' in \emph{CVPR}, 2014.

\bibitem{Gao2014}
J.~Gao, H.~Ling, W.~Hu, and J.~Xing, ``{Transfer Learning Based Visual Tracking
  with Gaussian Processes Regression},'' in \emph{ECCV}, 2014.

\bibitem{Zhang2014g}
J.~Zhang, S.~Ma, and S.~Sclaroff, ``{MEEM: Robust Tracking via Multiple Experts
  Using Entropy Minimization},'' in \emph{ECCV}, 2014.

\bibitem{Henriques2012}
J.~Henriques, R.~Caseiro, P.~Martins, and J.~Batista, ``{Exploiting the
  Circulant Structure of Tracking-by-Detection with Kernels},'' in \emph{ECCV},
  2012.

\bibitem{Hare2011}
S.~Hare, A.~Saffari, and P.~Torr, ``{Struck: Structured output tracking with
  kernels},'' in \emph{ICCV}, 2011.

\bibitem{Danelljan2017cvpr}
M.~Danelljan, G.~Bhat, F.~S. Khan, and M.~Felsberg, ``{ECO: Efficient
  Convolution Operators for Tracking},'' in \emph{CVPR}, 2017.

\bibitem{Danelljan2015iccv}
M.~Danelljan, H.~Gustav, F.~S. Khan, and M.~Felsberg, ``{Learning Spatially
  Regularized Correlation Filters for Visual Tracking},'' in \emph{ICCV}, 2015.

\end{thebibliography}
\end{document}